\documentclass[journal]{IEEEtran}
\usepackage{cite}
\ifCLASSINFOpdf
\else
\fi
\usepackage{amsmath}
\usepackage{url}
\usepackage{amsmath}
\usepackage{amsfonts}
\usepackage{amssymb}
\usepackage{float}
\ifCLASSOPTIONcompsoc
\usepackage[caption=false,font=normalsize,labelfont=sf,textfont=sf]{subfig}
\else
\usepackage[caption=false,font=footnotesize]{subfig}
\fi

\usepackage{tikz}
\usepackage{smartdiagram}
\usepackage{graphicx}
\usepackage{multirow}

\usepackage{color, colortbl}
\usepackage{xcolor}
\definecolor{lightgray}{gray}{0.1}
\definecolor{orange}{rgb}{1,0.5,0}
\begin{document}
%\title{Learning Driving Patterns  Using Bayesian Nonparametric Approaches}
%\title{Learning Driving Patterns Using Bayesian Nonparametric Approaches for Driving \\ Style Analysis}
\title{Driving Style Analysis Using Primitive Driving Patterns With Bayesian Nonparametric \\ Approaches}
%\title{Recognition of Driving Styles in Car-Following Behaviors Using Bayesian Nonparametric Approaches}

\author{Wenshuo~Wang,~\IEEEmembership{Student Member,~IEEE,}
        Junqiang~Xi,~%\IEEEmembership{Fellow,~OSA,}
        and~Ding~Zhao%,~\IEEEmembership{Life~Fellow,~IEEE}% <-this % stops a space
\thanks{This work was supported by The China Scholarship Council (CSC). (\textit{Corresponding Author: Junqiang Xi and Ding Zhao})}
\thanks{W. Wang is with the Department of Mechanical Engineering, Beijing Institute of Technology, Beijing, China,  100081, and also with Department of Mechanical Engineering, University of California, Berkeley, Berkeley, CA, 94720, USA. e-mail: wwsbit@gmail.com}% <-this % stops a space
\thanks{D. Zhao is with the Department of Mechanical Engineering, University of Michigan, Ann Arbor, MI, 48109, USA. e-mail: zhaoding@umich.edu}% <-this % stops a space
\thanks{J. Xi is with the Department of Mechanical Engineering, Beijing Institute of Technology, Beijing, China,  100081. e-mail: xijunqiang@bit.edu.cn}
}


\markboth{IEEE Transactions on Intelligent Transportation Systems,~Vol.~, No.~, July~2017}%
{Shell \MakeLowercase{\textit{et al.}}: Bare Demo of IEEEtran.cls for IEEE Journals}



\maketitle

\begin{abstract}
Analysis and recognition of driving styles are profoundly important to intelligent transportation and vehicle calibration. This paper presents a novel driving style analysis framework using the primitive driving patterns learned from naturalistic driving data. In order to achieve this, first, a Bayesian nonparametric learning method based on a hidden semi-Markov model (HSMM) is introduced to extract primitive driving patterns from time series driving data without prior knowledge of the number of these patterns. In the Bayesian nonparametric approach, we utilize a hierarchical Dirichlet process (HDP) instead of learning the unknown number of smooth dynamical modes of HSMM, thus generating the primitive driving patterns. Each primitive pattern is clustered and then labeled using behavioral semantics according to drivers' physical and psychological perception thresholds. For each driver, 75 primitive driving patterns in car-following scenarios are learned and semantically labeled. In order to show the HDP-HSMM's utility to learn primitive driving patterns, other two Bayesian nonparametric approaches, HDP-HMM and sticky HDP-HMM, are compared. The naturalistic driving data of 18 drivers were collected from the University of Michigan Safety Pilot Model Deployment (SPDM) database. The individual driving styles are discussed according to distribution characteristics of the learned primitive driving patterns and also the difference in driving styles among drivers are evaluated using the Kullback-Leibler divergence. The experiment results demonstrate that the proposed primitive pattern-based method can allow one to semantically understand driver behaviors and driving styles.
\end{abstract}


% \fi
\IEEEpeerreviewmaketitle

\section{Introduction}
%\subsection{Motivations}

\IEEEPARstart{D}{riving} style has great impact on eco-driving\cite{di2014stochastic}, road safety\cite{sagberg2015review}, and intelligent vehicles\cite{martinez17driving}. \textit{Driving style}, in this paper, refers to a set of dynamic activities/steps that a driver uses when driving, according to his/her personal judgment, experience and skills\cite{rafael2006impact,murphey2009driver}. Lots of previous research has focused on characterizing and analyzing driving styles. Most of them directly utilized the statistical metrics of measured driving data to feature driving styles. For example, the means, standard deviations, and maximums of brake pressure and throttle position were used to classify drivers into mild, moderate, and aggressive types in \cite{xu2015establishing,shi2015evaluating,wang2017driving}. Vaitkus, \textit{et al}. \cite{vaitkus2014driving} selected the mean, median, and standard deviation of longitudinal and lateral acceleration to describe driving styles. This kind of the above-mentioned approaches is easy to capture the static driving habits from a statistical perspective, but could not describe the dynamic process of drivers' behavioral semantics or decision making. 

Decomposing complex driver behaviors into simple, smaller and primitive patterns can facilitate identification and analysis of driving styles\cite{higgs2015segmentation}. \textit{Primitive driving pattern}, as generally defined in this paper, is the primitive segments that can be viewed as the basic composition of driver behaviors.  For example, drivers' car-following behaviors can be roughly segmented into three primitive patterns, i.e., closing in, keeping, and falling behind.  Driving behavior segmentation enables long sequences of observed driver behavior data to be segmented into smaller components, to formulate and gain insight into the driver's dynamic decision-making process \cite{okuda2013modeling,sekizawa2007modeling} and driving style \cite{higgs2015segmentation}. Driving pattern definition is related to how driving patterns are characterized, allowing a human or algorithmic observer to identify driving patterns from measured data. These definitions are often subjective,  application-driven, algorithmic-dependent, and tend to group into three categories:

\subsubsection{Physical Boundaries}
The definition of a pattern typically refers to a physical change that occurs when a driver's operation/decision starts or ends. These natural physical boundaries can be specified by changes of vehicle steering angle, brake/accelerator pedal position or their combination\cite{johnson2011driving}. These domain knowledge characteristics may be specific to a particular operation (e.g., turn left or turn right) or could generalize multiple operations (e.g., acceleration and lane change). For example, MacAdam, \textit{et al}. \cite{macadam1998using} manually classified car-following behavior into five patterns (i.e., closing in rapidly, close in, following, falling behind, and falling behind rapidly) based on the predefined thresholds of relative speed and relative distance, and then trained a neural network classifier to evaluate drivers' driving styles. Researchers also segmented lane-changing behavior into two patterns \cite{do2017human}  (i.e., longitudinal control pattern and lateral control pattern), three patterns \cite{nilsson2016if} (i.e., if, when, and how to perform lane change maneuvers), and five patterns \cite{nilsson2016lane} depending on vehicle position in the lane as well as with respect to surrounding vehicles. In general, the physical boundaries are empirically set according to suit requirements, which are more subjective.

\subsubsection{Template Boundaries}
Driving pattern can be defined by a user-provided template. One of the most popular template-based algorithms is the dynamic time warping, for example, applied to car-following behavior analysis\cite{taylor2015method}.  Defining a primitive pattern by a template or a set of template allows maximizing flexibility for users depending on special requirements. However, this approach is usually time-consuming for preparing the templates and may miss some special and meaningful templates. 

\subsubsection{Derived Metrics Boundaries} A primitive pattern can also be defined by a change in metrics (e.g., variance) or derived signals (e.g., hidden Markov model state transitions) based on supervised and unsupervised approaches. For example, Ma and Andreasson \cite{ma2007behavior} directly implemented a consolidated fuzzy clustering algorithm to classify different car-following regimes and then defined five car-following patterns, i.e., acceleration, stable following, braking, approaching, and opening. Higgs and Abbas \cite{higgs2015segmentation} developed a two-step algorithm to segment drivers' car-following behaviors based on eight predefined state-action variables (longitudinal acceleration, lateral acceleration, yaw rate, vehicle speed, lane offset, yaw angle, range, and range rate). After that, the method in \cite{higgs2015segmentation} resulted in 30 state-action clusters corresponding to driving patterns. These aforementioned approaches, however, require prior knowledge about the number of patterns or clusters.

\begin{figure}[t]
	\centering
	\includegraphics[width = 0.48\textwidth]{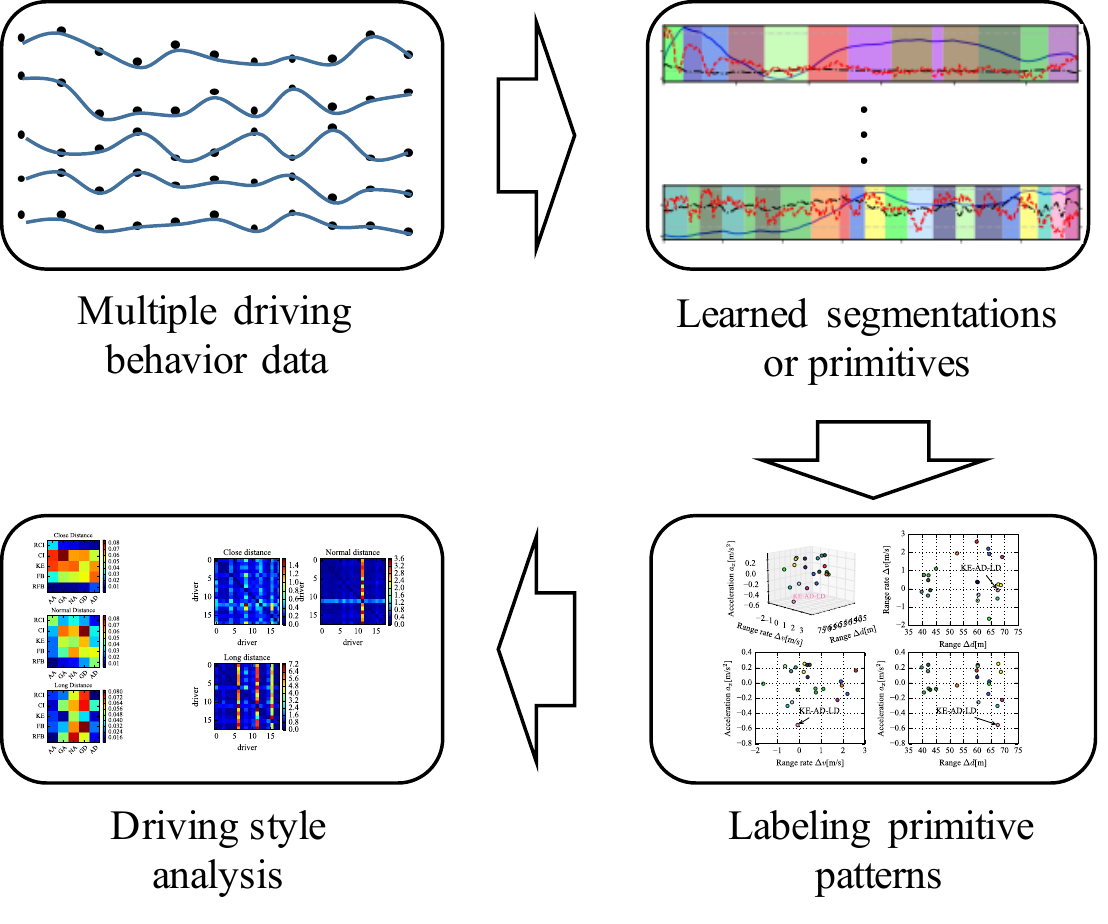}
	\caption{The proposed framework to analyze driving styles based on driving primitive patterns.}
	\label{figure:propose_method}
\end{figure}

Differing from previous research selecting the statistical features (e.g., mean and standard deviations) of the measured driving data and/or manually defining the number of patterns to characterize driving styles, in this paper, we proposed a learning-based framework (Fig. \ref{figure:propose_method}) for driving style analysis based on primitive driving patterns. The primitive driving pattern can reflect not only drivers' driving preferences but also the dynamic process of behavioral semantics and decision-making.  In our approach, rather than subjectively relying on hand-tuned models or predefined rules to select primitive driving patterns like in \cite{higgs2015segmentation,macadam1998using,ma2007behavior}, we introduced a Bayesian nonparametric approach to directly learn primitive driving patterns from time series driving data without requiring prior knowledge about the number of primitive patterns. For this purpose, a hierarchical modeling structure is used to learn the number of primitive driving patterns based on a hierarchical Dirichlet process (HDP) and a hidden semi-Markov model (HSMM). The primitive driving patterns are then labeled semantically for each driver according to drivers' physical and psychological perception thresholds with a K-mean clustering method. Finally, we use the distribution of primitive patterns to semantically analyze individual's driving styles and also use entropy index to illustrate the differences among drivers.

The remainder of this paper is organized as follows. Section II introduces the Bayesian nonparametric approach based on a hidden Markov model (HMM) and its extensions. Section III presents driving data collection and preprocessing. Section IV shows the experimental results of primitive driving patterns using different approaches. Section V analyzes the results of driving styles. Lastly, the conclusions are made in Section VI.

\section{Bayesian Nonparametric Learning Approaches Based on HMM}
Bayesian nonparametric learning methods have shown their powerful ability to model and predict driver behaviors \cite{hamada2016modeling,taniguchi2016sequence,taniguchi2015unsupervised} in the case where the number of primitive driving patterns is not exactly known. In this section, we illustrate a Bayesian nonparametric approach for learning driving patterns, i.e., HDP-HSMM. In order to understand this approach easily, we first show and discuss some basic concepts, including HMM, HSMM, HDP, HDP-HMM, and sticky HDP-HMM.

\subsection{HMM-Based Approaches}

\begin{figure}[t]
	\centering
	\subfloat[Hidden Markov model]{\includegraphics[width = 0.3\textwidth]{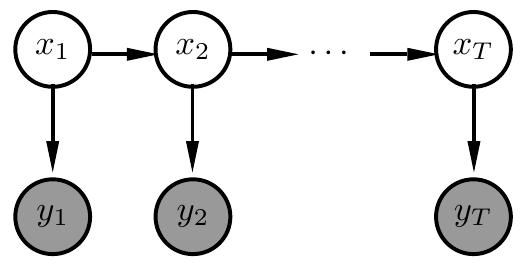}}\\
	\subfloat[Hidden semi-Markov model]{\includegraphics[width = 0.48\textwidth]{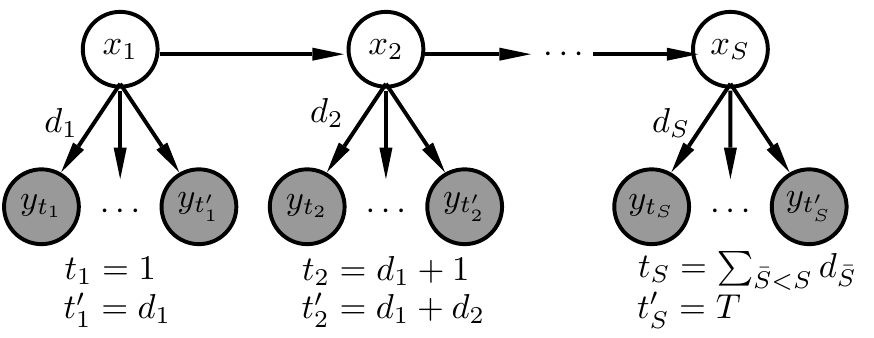}}
	\caption{A graphical model of HMM and HSMM. Shaded nodes are the observable states and unshaded nodes are the latent modes.}
	\label{fig:HMM-and-HSMM}
\end{figure}

\begin{figure*}[t]
	\centering
	\subfloat[HDP-HMM]{\includegraphics[width = 0.27\textwidth]{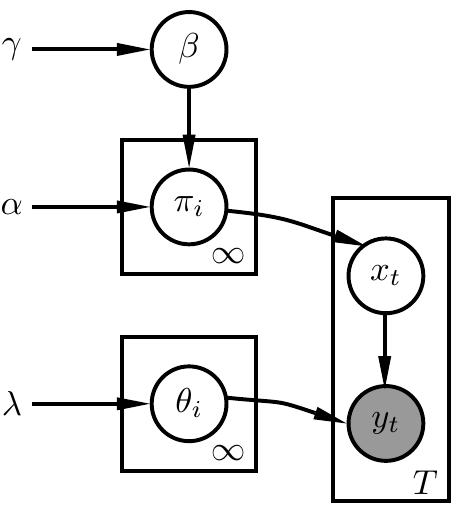}}~
	\subfloat[sticky HDP-HMM]{\includegraphics[width = 0.27\textwidth]{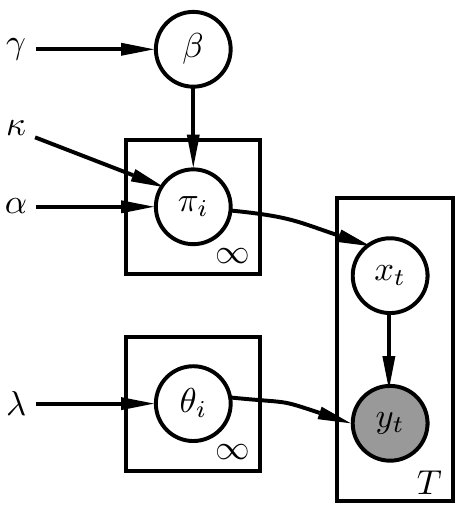}}~
	\subfloat[HDP-HSMM]{\includegraphics[width = 0.37\textwidth]{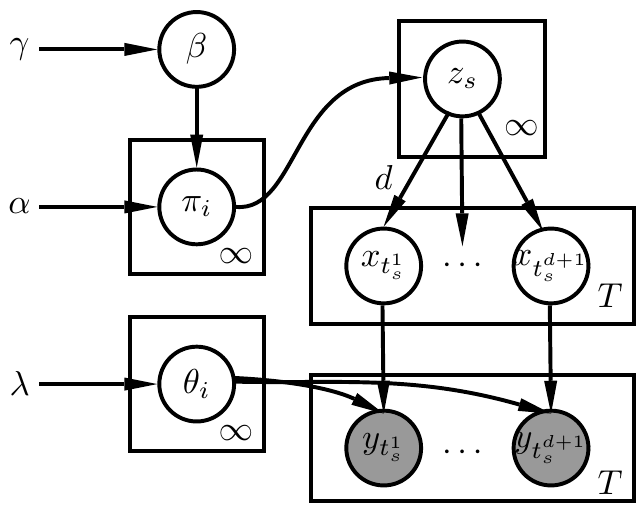}}
	\caption{Graphical models of three Bayesian nonparametric HMM-based approaches for an univariate time series data with a length $ T $.}
	\label{fig:Model_overview}
\end{figure*}

\subsubsection{Hidden Markov Model} In this work, we view the dynamic process of primitive driving patterns in driving behaviors as a Markov process. Thus, driver behaviors can be modeled based on the structure of HMM. The core of HMM consists of two layers: a layer of hidden \textit{state} and a layer of \textit{observation} or \textit{emission}, as shown in Fig. \ref{fig:HMM-and-HSMM}(a), where the shaded nodes are observations, $ y_{t} $, and the unshaded nodes are latent states $ x_{t} $, i.e., primitive driving patterns in this work. The hidden state sequence, $ \boldsymbol{x} = \{x_{t}\}_{t = 1}^{T}$, is a sequence of primitive driving patterns over a finite set $ \mathcal{X} $, i.e., $ x_{t} \in \mathcal{X} $. The transition probability from primitive driving patterns $ i $ to $ j $ is noted as $ \pi_{i,j}  = p(x_{t+1} = j| x_{t} = i)$ and the transition matrix between these patterns is $ \boldsymbol{\pi} = \{\pi_{i,j}\}_{i,j = 1}^{|\mathcal{X}|} $, where $ |\mathcal{X}| $ indicates the size of pattern set $ \mathcal{X} $. The distribution of observations or emissions $ y_{t} $ given current hidden states is defined by $ p(y_{t}|x_{t}, \theta_{i}) $, where $ \theta_{i} $ is the emission parameter for mode $ i $. Thus, we can describe HMM by
\begin{subequations}\label{equation:HMM}
	\begin{align}
	x_{t}|x_{t-1}  &\sim \pi_{x_{t-1}} \\
	y_{t}|x_{t} & \sim F(\theta_{x_{t}})
	\end{align}
\end{subequations}
where $ F(\cdot) $ is an indexed family of distribution. %, $ e_{t}^{x_{t]}} \sim \mathcal{N}(\mu_{0},\Sigma_{0})$ is the observation noise.

\subsubsection{Hidden Semi-Markov Model}
The hidden semi-Markov model (HSMM), as an extension of HMM, is traditionally defined by allowing the underlying process to be a semi-Markov chain, which means that each state has a variable duration \cite{yu2010hidden}, as shown in Fig. \ref{fig:HMM-and-HSMM}(b). Several approaches can be used to define HSMM depending on the assumptions and applications. In this paper, we assume that each hidden state's duration is given over an explicit distribution, also called \textit{explicit duration HMM} \cite{rabiner1989tutorial}. Therefore, as defined in \cite{johnson2013bayesian}, we augment the generative process of a standard HMM with a random state duration time, drawn from some state-specific distribution when the state is entered. Here, we use the random variable $ d_{t} $ to denote the duration of a state that enters at time $ t $, and $ p(d_{t}|x_{t} = i) $ denotes the probability mass function for $ d_{t} $. Similar to HMM, we can define HSMM by
\begin{subequations}\label{equation:HSMM}
	\begin{align}
	x_{s}|x_{s-1}&\sim \pi_{x_{s-1}} \\
	 d_{s}&\sim g(\omega_{s}) \\
	y_{t}|x_{s},d_{s}&\sim F(\theta_{x_{s}}, d_{s})
	\end{align}
\end{subequations}
where $ g(\omega_{s}) $ is a state-specific distribution over state duration $ d_{s} $.

\subsubsection{Hierarchical Dirichlet Process}
As aforementioned in Section I, we assume that the number of latent dynamic modes or patterns in (\ref{equation:HMM}) and (\ref{equation:HSMM}) is priorly unknown and these modes of HMM and HSMM are subject to a specific distribution defined over a measure space. 

The \textit{Dirichlet process} (DP) is a measure on measures \cite{teh2006hierarchical}, denoted by DP($ \gamma, H $), and provides a distribution over discrete probability measures with an infinite collection of atoms \cite{fox2011bayesian}

\begin{subequations}
	\begin{align}
	G_{0}& = \sum_{i=1}^{\infty}\beta_{i}\delta_{\theta_{i}}, \ \ \  \theta_{i} \sim H\\
	\beta_{i}& = \nu_{i}\prod_{\ell=1}^{i-1}(1-\nu_{\ell}), \ \ \ \nu_{i} \sim \mathrm{Beta}(1,\gamma)
	\end{align}
\end{subequations}
 on a parameter space $ \Theta $ that is endowed with a base measure $ H $. Here, the weights $ \beta_{i} $ are sampled by a stick-breaking construction \cite{teh2006hierarchical} and we denote $ \beta \sim \mathrm{GEM}(\gamma)$, with $ \beta = [\beta_{1}, \beta_{2}, \cdots] $.
 
According to the above discussion, a HDP \cite{teh2006hierarchical} is able to be used to define a prior on the set of HMM transition probability measures $ G_{j} $
\begin{equation}
G_{j} = \sum_{i=1}^{I}\pi_{ji}\delta_{\theta_{i}}
\end{equation}
where $ \delta_{\theta} $ is a mass concentrated at $ \theta $. Assuming that each discrete measure $ G_{j} $ is a variation on a global discrete measure $ G_{0} $, thus the Bayesian hierarchical specification takes $ G_{j}\sim \mathrm{DP}(\alpha, G_{0}) $,  where $ G_{0} $ draws from a DP($ \gamma, H $), i,e.,
\begin{subequations}\label{equation:DP}
	\begin{align}
	G_{0}& = \sum_{i=1}^{\infty}\beta_{i}\delta_{\theta_{i}}, \ \ \ \beta|\gamma \sim \mathrm{GEM}(\gamma) \\
	G_{j}& = \sum_{i=1}^{\infty}\pi_{ji}\delta_{\theta_{i}}, \ \ \ \pi_{j}|\alpha, \beta \sim \mathrm{DP}(\alpha, \beta) \\
	\theta_{i}|H&\sim H
	\end{align}
\end{subequations} 
where $ \pi_{j} = [\pi_{j1}, \pi_{j2}, \cdots] $.

\subsubsection{Sticky HDP-HMM and HDP-HSMM}
Based on the above discussion, by applying the HDP prior to the HMM and HSMM, we can obtain the HDP-HMM, sticky HDP-HMM \cite{teh2006hierarchical}, and HDP-HSMM \cite{johnson2013bayesian}, as shown in Fig. \ref{fig:Model_overview}.

For the sticky HDP-HMM($  \gamma,\alpha, H $), by adding an extra parameter $ \kappa >0 $ that biases the process toward self-transition in (\ref{equation:DP}b), increasing the expected probability of self-transition by an amount proportional to $ \kappa $\cite{fox2011bayesian}, we can obtain
	\begin{subequations}\label{equation:sticky_HDP_HMM}
		\begin{align}
		\beta|\gamma&\sim \mathrm{GEM}(\gamma) \\
		\pi_{i}|\alpha,\beta,\kappa&\sim \mathrm{DP}(\alpha + \kappa, \frac{\alpha\beta + \kappa \delta_{i}}{\alpha + \kappa}), \ i = 1, 2, \cdots \\
		x_{t}|x_{t-1}&\sim \pi_{x_{t-1}}, \ t = 1,2, \cdots, T  \\ 
		y_{t}|x_{t},\theta_{x_{t}}&\sim F(\theta_{x_{t}}), \ t = 1,2, \cdots, T \\
		\theta_{i}|H&\sim  H, \ i = 1,2, \cdots.
		\end{align}
	\end{subequations}
where $ T $ is the data length. Note that when $ \kappa = 0 $ in (\ref{equation:sticky_HDP_HMM}b), the original HDP-HMM is obtained.
	
Similarly, for the HDP-HSMM($ \gamma, \alpha, H, G $), we can describe it using \cite{johnson2013bayesian}
	\begin{subequations}
		\begin{align}
		\beta|\gamma&\sim \mathrm{GEM}(\gamma) \\
		\pi_{i}|\alpha, \beta & \sim \mathrm{DP}(\alpha, \beta), \ \ \ i = 1,2, \cdots \\
		(\theta_{i}, \omega_{i}) & \sim H \times G, \ \ \ i = 1,2, \cdots  \\
		z_{s} & \sim \bar{\pi}_{z_{s-1}}, \ \ \ s = 1,2, \cdots \\
		d_{s} & \sim g(\omega_{z_{s}}), \ \ \ s = 1,2, \cdots \\
		x_{t_{s}^{1}:t_{s}^{d_s+1}} & = z_{s} \\
		y_{t_{s}^{1}:t_{s}^{d_s+1}} & = F(\theta_{x_{t}})
		\end{align}
	\end{subequations}
	where $ \bar{\pi} = \frac{\pi_{ij}}{1-\pi_{ii}} (1-\delta_{ij})$ is used to estimate self-transition in the state sequence $ z_{s} $.

%\subsection{Structures for a Single Event Sequence}

\subsection{Observation (or Emission) Model}
The observation model is determined by the type of function $ F(\theta_{i}) $, which can be Gaussian emissions \cite{mahboubi2016learning} or switch linear dynamic models (SLDSs)\cite{fox2011bayesian} (e.g., vector autoregressive). One main challenge with non-parametric approaches is that one must derive all the necessary expressions to properly perform inference\cite{ryden2008versus}. Here, in order to make our algorithm tractable, we assume that observations are drawn from a Gaussian distribution like in \cite{mahboubi2016learning}. When observations are assumed to be drawn from a Gaussian distribution, the $ \theta_{i} $ can be set as $ \theta_{i} = [\mu_{i}, \Sigma_{i}] $. Therefore, if the priors for observations and transition distributions are learned correctly, the full-conditional posteriors can be computed using Gibbs sampling method. Johnson and Willsky \cite{johnson2013bayesian} present further details of the inference method using Gibbs sampling methods. % or from a SLDS, then $  \theta_{i} = [\boldsymbol{A}_{i}, \Sigma_{i}]$.

\begin{table}[t]
	\centering
	\caption{\textsc{Parameter Values for Models}}
	\begin{tabular}{ccc}
		\hline
		\hline
		parameter & description  & value\\
		\hline
		$ (a_{\alpha}, b_{\alpha}) $ & $ \alpha $ gamma prior  & (1,1) \\
		$(a_{\gamma}, b_{\gamma})$ & $ \gamma $ gamma prior & (1,1) \\
		$ (a_{\kappa}, b_{\kappa}) $ & $ \kappa $ gamma prior& (100,1)\\
		%		$ \mu_{i} $ & $ \theta_{i} $ prior mean & 0\\
		%		$ \Sigma_{i}$ &  $ \theta_{i} $ prior covariance & $ I_{1\times 1} $\\
		$ n_{0} $ & IW prior degree of freedom & $ N+2 $ \\
		$ S_{0} $ & IW prior scale & 0.75$\cdot \bar{\Sigma} $\\
		\hline
		\hline
	\end{tabular}
	\label{Table:parameters}
\end{table}

\subsection{Learning Procedure}
We develop and test the developed models based on Johnson and Willsky's \texttt{pyhsmm}\footnote[1]{\url{https://github.com/mattjj/pyhsmm}} module \cite{johnson2013bayesian} and Fox's code \footnote[2]{\url{https://www.stat.washington.edu/~ebfox/software.html}}\cite{wulsin2014modeling}. In this work, the hyperparameters are determined using following rules: 
\begin{enumerate}
	\item We place a Gamma($ a, b $) prior on the hyperparameters $ \gamma $, $ \alpha $, and $ \kappa$\cite{hamada2016modeling,fox2011bayesian}, as shown in Table \ref{Table:parameters}, where $ N $ is the dimension of input data.
	\item The Inverse-Wishart (IW) prior is conjugate to the Gaussian distributions \cite{wang2017learning} and SLDS parameter set \cite{fox2011bayesian}, thus the hyperparameters for $ \theta_{i} $ are taken to be from an IW with a hyper-parameter $ \gamma $, that is,
	
	\begin{equation}
	\Sigma_{i}|n_{0}, S_{0} \sim \mathrm{IW} (n_{0}, S_{0})
	\end{equation}
	where $  n_0 $ is IW prior degree of freedom, $ S_{0} $ is the IW prior scale, and $S_{0} = 0.75 \bar{\Sigma} $, where $ \bar{\Sigma} $ is the covariance of the observed data. 
\end{enumerate}
In this work, the observation variables are generated from a Gaussian model and we set $ \mu_{i} = 0 $ according to \cite{hamada2016modeling}.
 % when the observation variables are emitted from a SLDS model, the computation of $ \boldsymbol{A}_{i} $ refers to \cite{fox2011bayesian}.

\begin{figure}[t]
	\centering
	\includegraphics[width = \linewidth]{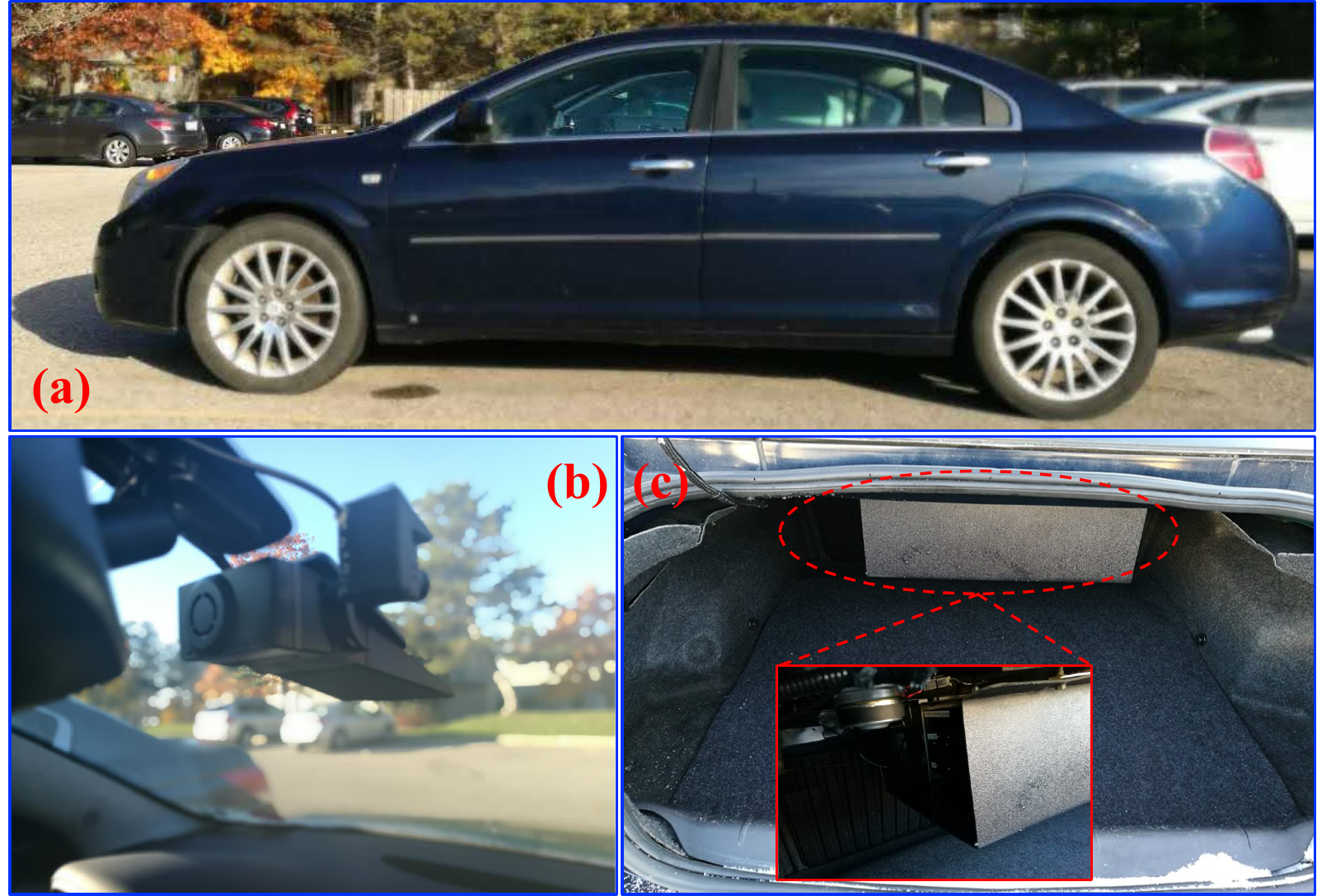}
	\caption{Example of equipment for data collection. (a) Vehicle; (b) Mobileye; and (c) Data acquisition systems.}
	\label{figure:experiment_equipment}
\end{figure}

\section{Car-Following Data Collection and Preprocessing}
In this work, we apply the developed methods to analyze driving styles regarding car-following behaviors.  The driving scenarios and data collection and processing are presented as follows.

\subsection{Equipment and Participants}
All driving data we used were extracted from the Safety Pilot Model Deployment (SPMD) database, which recorded the naturalistic driving of 2,842 equipment vehicles in Ann Arbor, Michigan, for more than two years. We used 18 equipped vehicles (i.e., 18 drivers) to run experiments and collected on-road data. The experiment vehicles were equipped with data acquisition systems and Mobieye, as shown in Fig. \ref{figure:experiment_equipment}. The road information (e.g., lane width, lane curvature) and the surrounding vehicle's information (e.g., relative distance, relative speed) were recorded by Mobileye. The subject vehicle information such as speed, steering angle, acceleration/brake pedal position were extracted from CAN-bus signal. All of the data were recorded at 10 Hz.

Drivers had an opportunity to become accustomed to the equipped vehicles. They performed casual daily trips for several months without any restrictions on or requirements for their trips, the duration of the trips, or their driving style. The data processing and recording equipment were hidden from the drivers, thus avoiding the influence of recorded data on driver behavior.

\begin{figure}[t]
	\centering
	\includegraphics[width = 0.48\textwidth]{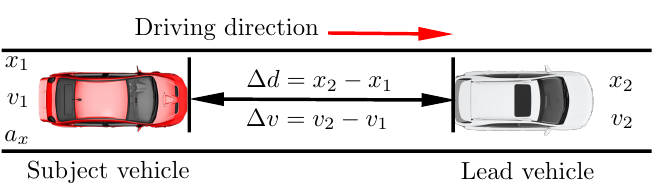}
	\caption{Car-following scenarios}
	\label{figure:carfollowing}
\end{figure}

\begin{figure}[t]
	\centering
	\includegraphics[width = 0.48\textwidth]{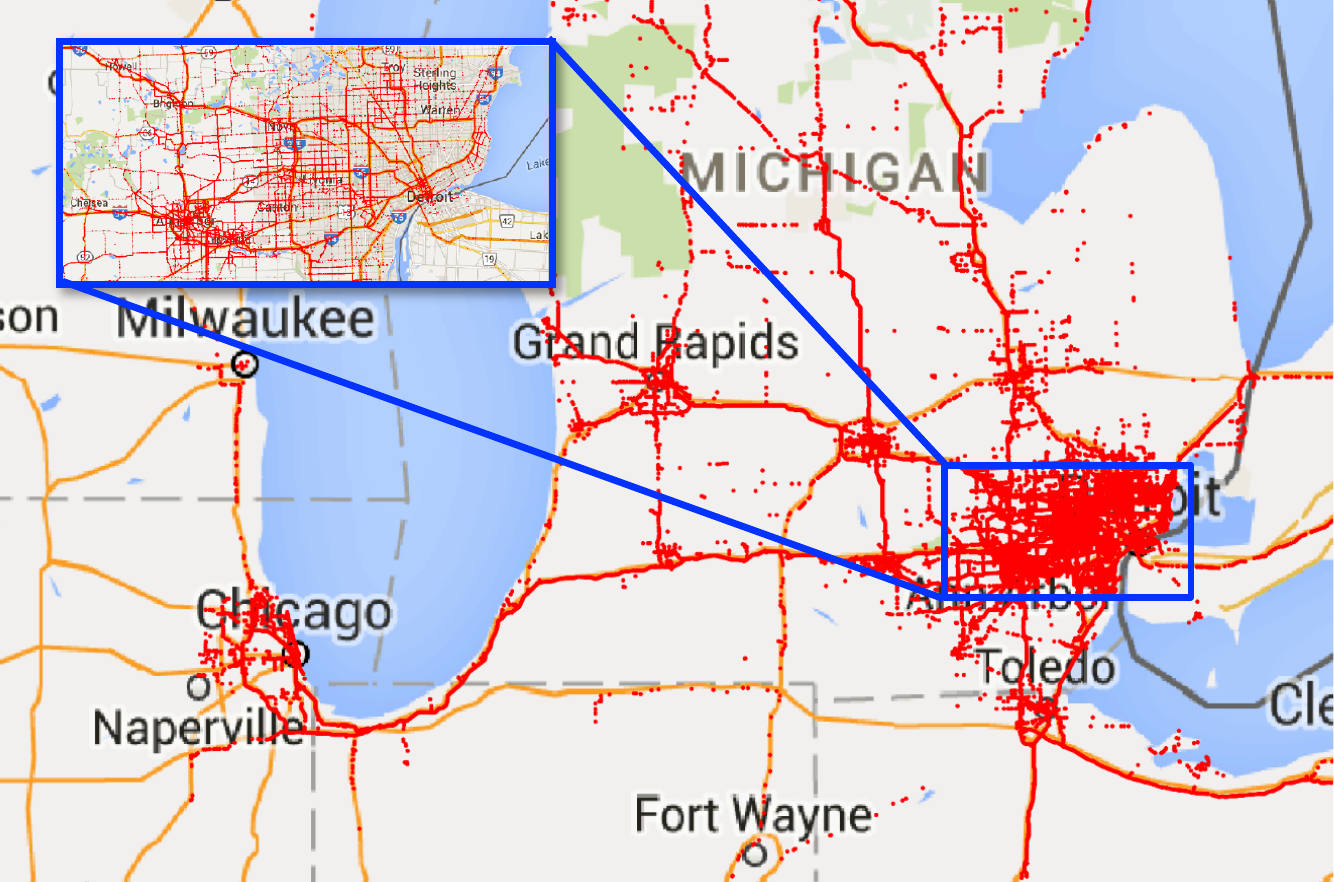}
	\caption{Trajectories of all car-following events.}
	\label{figure:trajectory}
\end{figure}

\begin{figure*}[t]
	\centering
	\subfloat[]{\includegraphics[width = 0.5\textwidth]{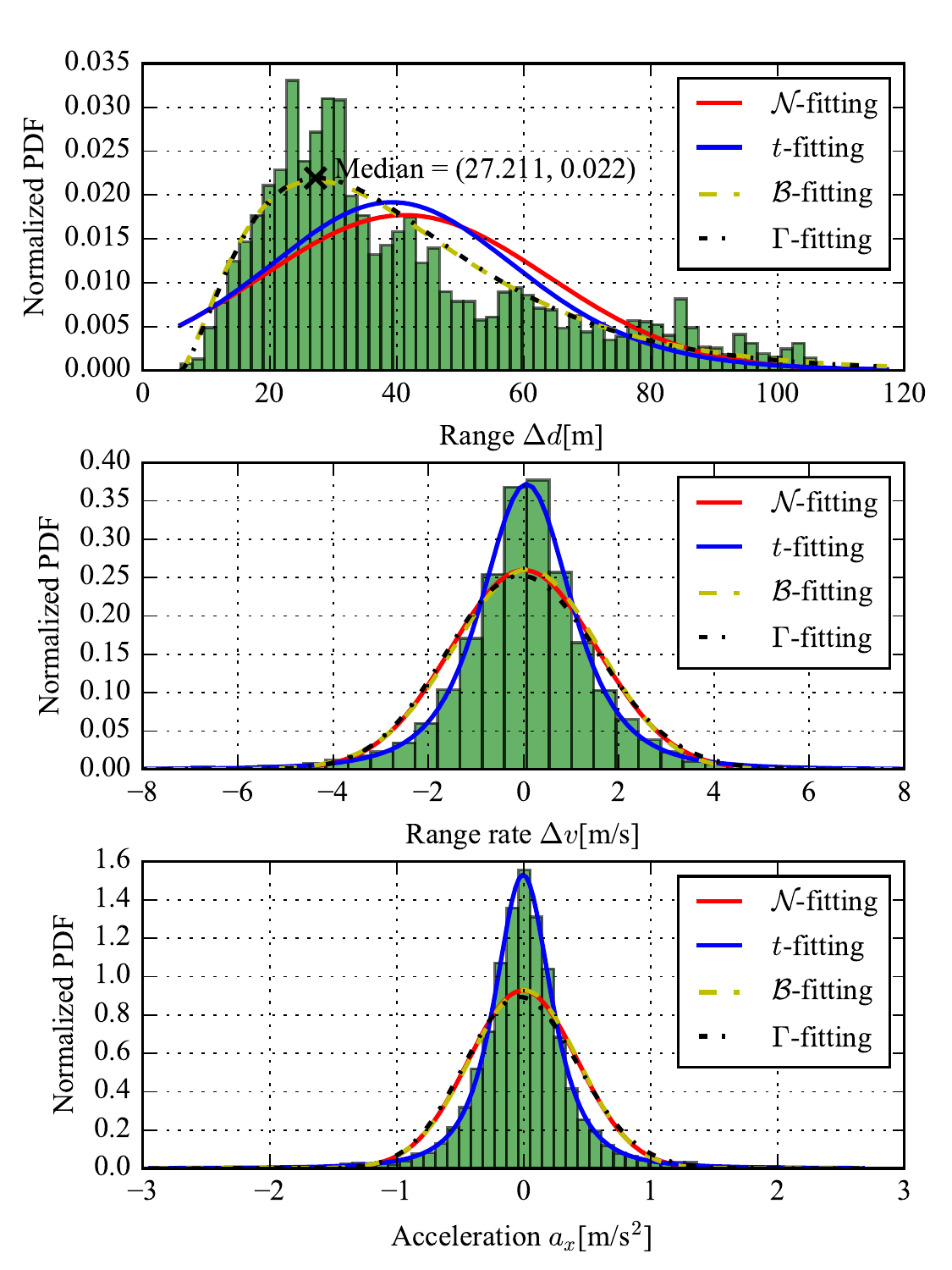}}
	\subfloat[]{\includegraphics[width = 0.5\textwidth]{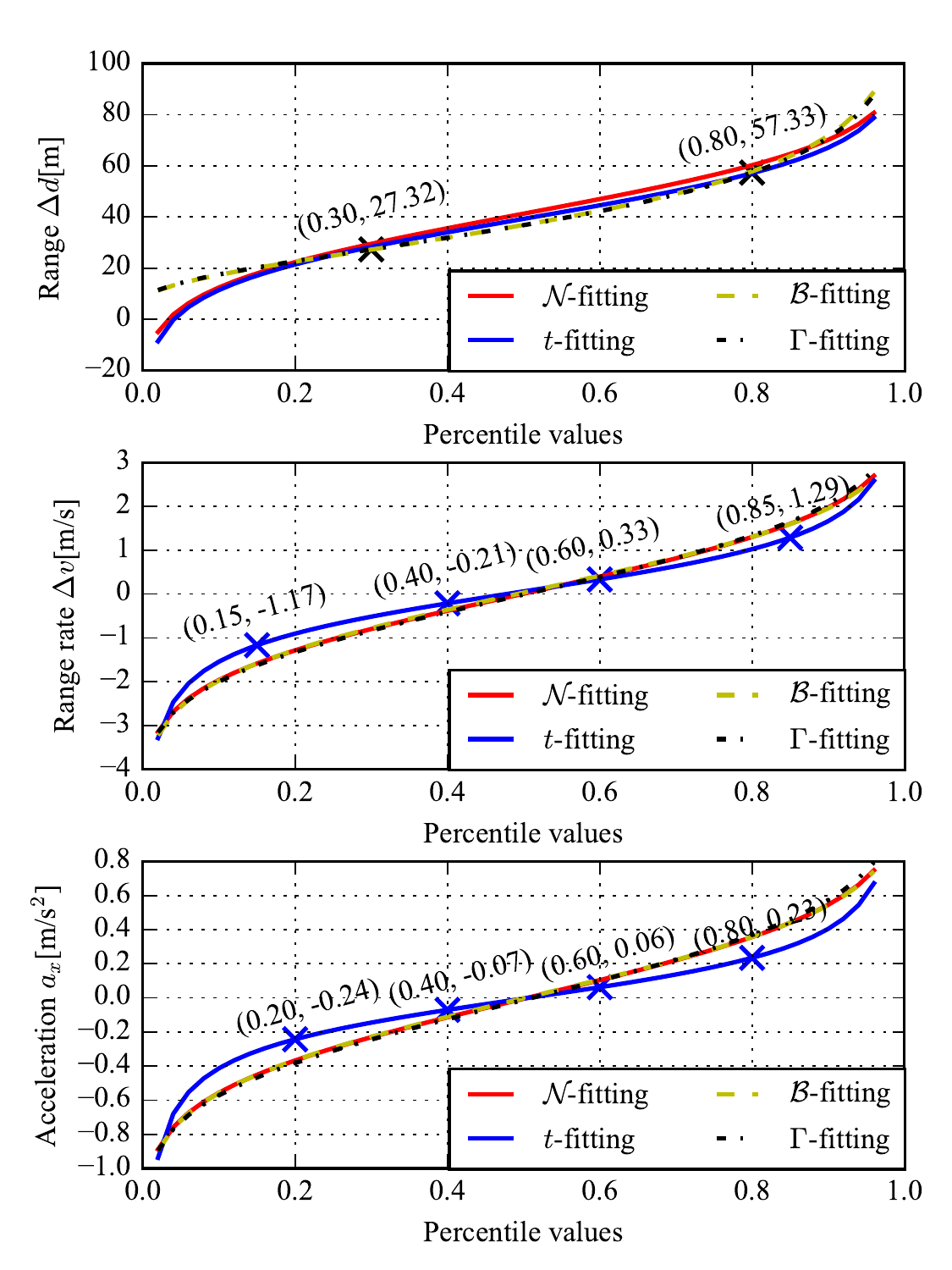}}
	\caption{(a) The statistical fitting results and (b) threshold values of three variables. Four approaches are used to fit the data, i.e., Normal ($ \mathcal{N} $) distribution, Beta ($ \mathcal{B} $) distribution, Student-$ t $ ($ t $) distribution, and Gamma ($ \Gamma $) distribution.}
	\label{figure:statis_thres}
\end{figure*}

\subsection{Data Extraction and Preprocessing}
In order to extract car-following events  (Fig. \ref{figure:carfollowing}) from the SPMD database and capture drivers' dynamic preference when driving as demonstrated in \cite{wang2017development}, we selected the following variables:
\begin{enumerate}
	\item The subject vehicle acceleration $ a_{x} $,  which can directly reflect driver's intents and driving preference.
	\item The relative range between subject and lead vehicles $ \Delta d $, which is computed by $ \Delta d = x_{2} - x_{1} $. It can reflect driver's preference in headway, where $ x_1 $ and $ x_{2} $ are the global positions of subject vehicle and lead vehicle, respectively.
	\item The relative range rate $ \Delta v $ which is computed by $ \Delta v = v_2 - v_1 $.  It can capture the dynamical relationship between two vehicles, where $ v_{1} $ and $ v_{2} $ are the velocities of subject vehicle and lead vehicle, respectively.
\end{enumerate}
Then, the car-following data consisting of the three feature variables were extracted from the SPMD database under the following conditions: (1) There was a lead vehicle at the same lane with the subject vehicle; (2) the lead distance was less than 120 m; (3) the vehicle speed was large than 18 km/h (i.e., 5 m/s); (4) if a surrounding vehicle was cutting in, then the car-following event ended; (5) the duration of each single car-following event was larger than 50 s\cite{wang2017much}. Fig. \ref{figure:trajectory} presents the trajectories of all car-following events in the SPMD database.

When training a model, in order to reduce the scale influence of different variables on the segmentation results, we additionally normalized each variable of observation vector for each event so that the empirical variance of the set was equal to one by

\begin{equation}
\bar{\boldsymbol{x}}^{(l)}_{m} = \frac{\boldsymbol{x}^{(l)}_{m}-\boldsymbol{\mu}_{m}}{\boldsymbol{\sigma}_{m}}, \ l = 1, 2, \cdots, L, \ m = 1,2,\cdots, M
\end{equation}
where $ \boldsymbol{x} = [\Delta d, \Delta v, a_{x}]^{\top} $, $ l $ is the number of event,  $ \boldsymbol{\mu}_{m} $ and $ \boldsymbol{\sigma}_{m} $ are the mean and covariance of all events for $ m^{\mathrm{th}} $ driver, with $ M $ = 18. Then we applied the  Bayesian nonparametric learning approaches presented in Section II to segment the normalized time series data, thus obtaining the regimes among primitive driving patterns.

\begin{table}[t]
	\centering
	\caption{\textsc{Variable Segmentation}}
	\label{Table:Var_Sem}
	\begin{tabular}{c|c|c}
		\hline
		\hline
		Variable	& Variable state           &   Threshold   \\ \hline
		\multirow{3}{*}{Range  [m]}  & Long distance (LD)  & $ > $ 57.33 \\ \cline{2-3} 
		& Normal distance (ND)          &     [27.32, 57.33]    \\ \cline{2-3} 
		& Close  distance (CD)           &    [5.00, 27.32]   \\ \hline
		\multirow{5}{*}{Range rate  [m/s]} & Rapidly closing in (RCI) &   $ < $ -0.90  \\ \cline{2-3} 
		& Closing in (CI)       &    [-1.17, -0.21]  \\ \cline{2-3} 
		& Keeping (KE)            &   [-0.21, 0.33] \\ \cline{2-3} 
		& Falling behind  (FB)  &  [0.33, 1.29]    \\ \cline{2-3} 
		& Rapidly falling behind (RFB)   &  $ > $ 1.29  \\ \hline
		\multirow{5}{*}{Acceleration [m/s$ ^{2} $]} & Aggressive acceleration (AA) & $ > $ 0.23 \\ \cline{2-3} 
		& Gentle acceleration (GA) &  [0.06, 0.23] \\ \cline{2-3} 
		& No acceleration (NA)      & [-0.07, 0.06] \\ \cline{2-3}
		& Gentle deceleration (GD)  & [-0.24, -0.07] \\ \cline{2-3} 
		& Aggressive deceleration (AD) & $ < $ -0.24 \\ \hline \hline
	\end{tabular}
\end{table}

\subsection{Variable Segmentation and Threshold Selection}
In order to easily make a semantic explanation for primitive driving patterns, we classify each variable into different levels based on drivers' physical and psychological perception thresholds corresponding to their statistical feature, as shown in Table \ref{Table:Var_Sem}. More specifically, we fit them using different distributions to determine the threshold of each variable from the statistical perspective.  

Fig. \ref{figure:statis_thres}(a) shows the fitting results of the relative range ($ \Delta d $), relative range rate ($ \Delta v $), and subject vehicle acceleration ($ a_{1} $) using four distributions, including normal distribution ($ \mathcal{N} $), Beta distribution ($ \mathcal{B} $), Student-$ t $ distribution ($ t $), and Gamma distribution ($ \Gamma $) \cite{bishop2007pattern}. We can see that: 1) for range rate and acceleration, the $ t $-distribution achieves a better fitting performance than other three distributions, and 2)  for the range, the $ \Gamma $-distribution and the $ \mathcal{B} $-distribution obtain a better fitting performance than other two distributions. Based on perceptible characteristics of variables and driver's comfortable thresholds, we select the percentile value of range with the $ \Gamma $-fitting results, and select the percentile values of range rate and acceleration with the $ t $-fitting results, as illustrated in Fig. \ref{figure:statis_thres}(b). The selection procedure is discussed as follows:

\begin{itemize}
	\item Relative range ($ \Delta d $): Relative range is divided into three levels: long distance (LD), normal distance (ND), and close distance (CD).  Note that from the top plot of Fig. \ref{figure:statis_thres}(a), the lower threshold of 27.32 m matches the median value 27.21 m and the upper threshold of 57.33 m is close to the threshold 61 m which is usually used to distinguish the free-flow regime \cite{ahmed1999modeling}. Therefore, the threshold 30 and 85 percentile values of headway distance are selected, corresponding to 27.32 m and 57.33 m, respectively .
	
	\item Relative range rate ($ \Delta v $): From \cite{yang2010development}, we know that drivers' velocity difference perception threshold is $ |\Delta v^{\mathrm{thr}}| \in$ [0.05, 0.2]  m/s. Therefore, it is conservative enough to assume that human driver hardly feel the change of range rate when $ |\Delta v| <$ 0.2 m/s. From \cite{fischer2012evaluation}, we also know that driver can distinctly feel the change of range rate when $ |\Delta v| >  $ 1.3 m/s. Therefore, the thresholds are selected the percentile values of 15, 40, 60, and 85 for range rate, corresponding to range rate values of -1.17 m/s, -0.21 m/s, 0.33 m/s, and 1.29 m/s, respectively, as shown in Table \ref{Table:Var_Sem} and the middle plot of Fig. \ref{figure:statis_thres}(b). Thus, the range rate is divided into five segments, similar to\cite{macadam1998using}: rapidly closing in (RCI), closing in (CI), keeping (KE), falling behind (FB), and rapidly falling behind (RFB).
	
	\item Acceleration ($ a_{x} $): Acceleration is divided into five levels based on drivers' vestibular and kinesthetic thresholds $ |a_{x}^{\mathrm{thr}}| \approx $ 0.05 m/s$ ^{2} $\cite{macadam2003understanding} and longitudinal acceleration comfort threshold  $ |a_{x}^{\mathrm{thr}}| \approx  $ 0.2 m/s$ ^{2} $ \cite{ioannou1994precursor}. In this work, we use the percentile values of 20, 40, 60, and 80 for acceleration as the thresholds, with -0.24 m/s$ ^{2} $, -0.07 m/s$ ^{2} $, 0.06 m/s$ ^{2} $, and 0.23 m/s$ ^{2} $, which quantitatively match the kinesthetic perception and comfort thresholds. Thus, the acceleration is segmented into aggressive acceleration (AA), gentle acceleration (GA), no acceleration (NA), gentle deceleration (GD), and aggressive deceleration (AD), as shown in Fig. \ref{figure:statis_thres}(b) and Table \ref{Table:Var_Sem}.
\end{itemize}

\begin{figure}[t]
	\centering
	\includegraphics[width = 0.48\textwidth]{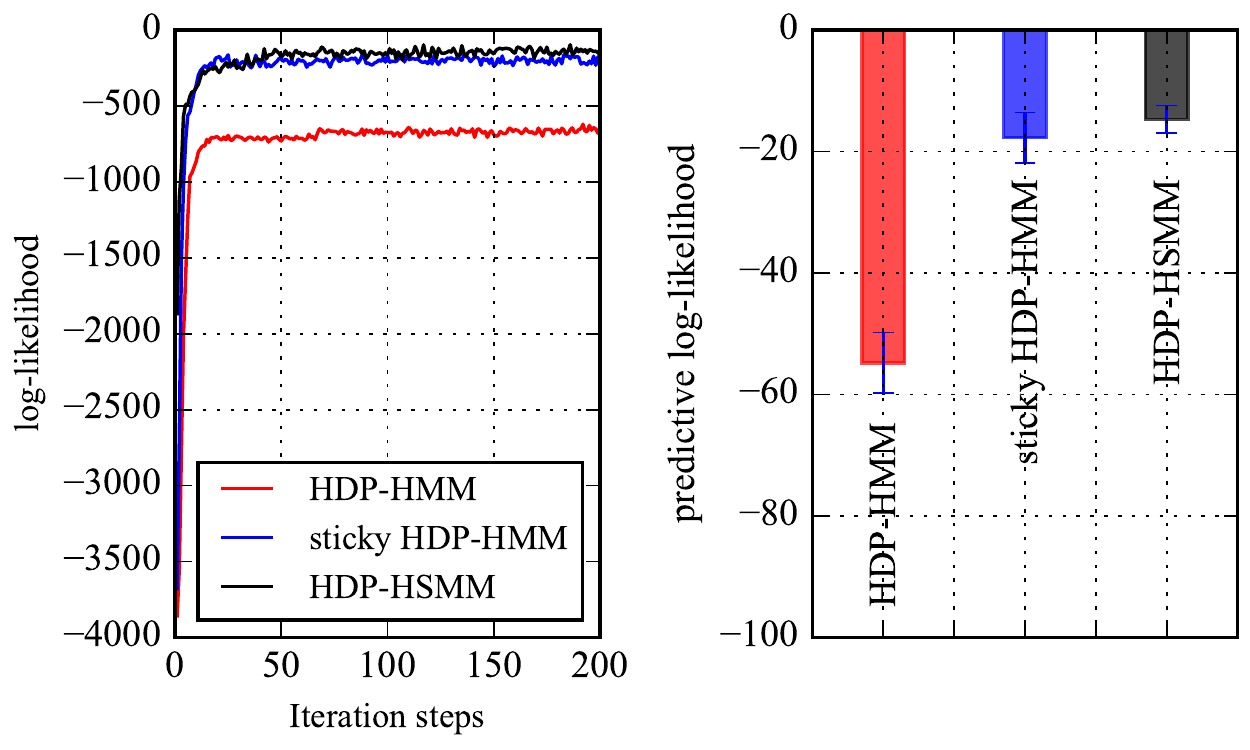}
	\caption{The log-likelihood of modeling the training data (left) and the predictive log-likelihood of test data (right) for all three methods.}
	\label{figure:Method_Comparison}
\end{figure}

\begin{figure*}[t]
	\centering
	\includegraphics[width = 0.85\textwidth]{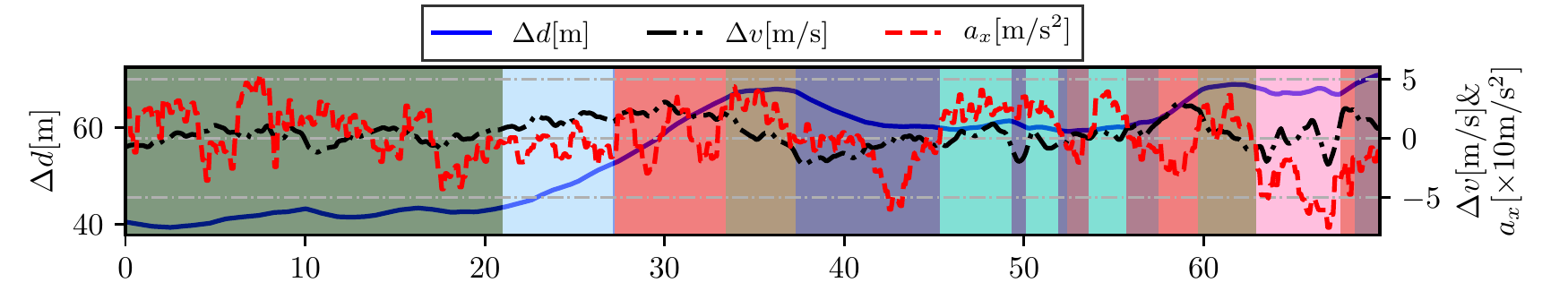} \\
	\includegraphics[width = 0.85\textwidth]{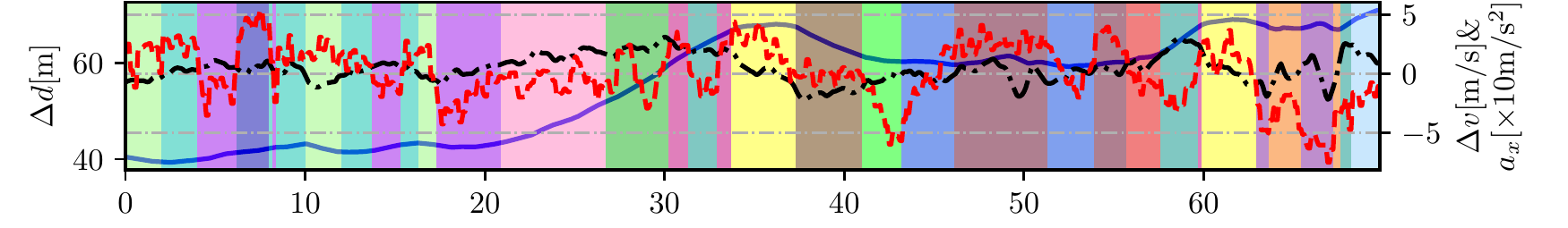} \\
	\includegraphics[width = 0.85\textwidth]{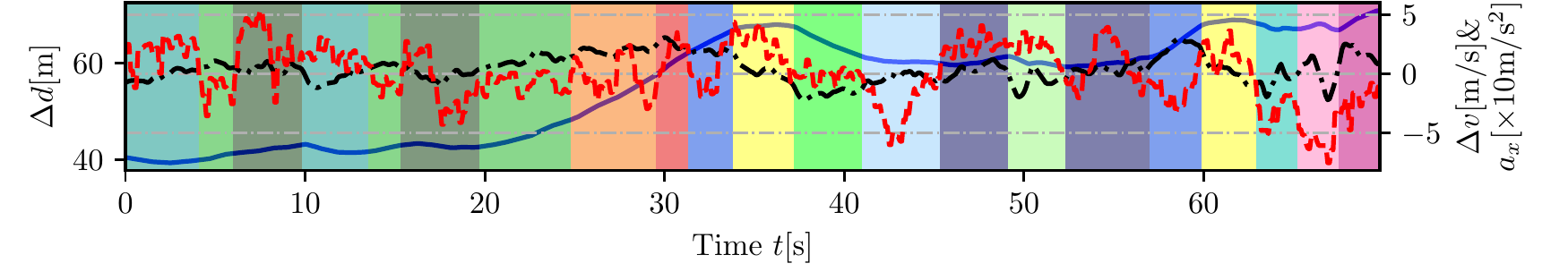}\\
	%	\subfloat[]{\includegraphics[width = 0.9\textwidth]{Figs/Res_HDP_HMM}}\\
	%	\subfloat[]{\includegraphics[width = 0.9\textwidth]{Figs/Res_StickyHDP_HMM}}\\
	%	\subfloat[]{\includegraphics[width = 0.9\textwidth]{Figs/Res_HDP_HSMM}}\\
	\caption{Example of experiment results of one event for driver \#0 using three different methods: HDP-HMM (top) with 9 patterns, sticky HDP-HMM (middle) with 17 patterns, and HDP-HSMM (bottom) with 14 patterns.}
	\label{figure:seg_results1}
\end{figure*}

Based on the aforementioned levels of each feature variable, we can obtain 75 different primitive car-following patterns ($ 3\times5\times5 = 75$) by combining the level of each feature variable (i.e., range, range rate, and acceleration) in Table \ref{Table:Var_Sem}. This allows us to label the primitive pattern of time series driving data in semantics, allowing one to analyze driving style and dynamic mode characteristics from a different perspective.

\section{Model Evaluation and Result Analysis}
In this section, the method utility for car-following behaviors and the learning results of the primitive driving patterns are discussed and analyzed.

\subsection{Method Performance Evaluation}
\subsubsection{Evaluation Method}
One standard approach to evaluate the model performance is using the leave-one-out cross-validation (LOO-CV) method with splitting data into training set and and test set. However, the ground truth of the segmentation results for test dataset is difficult to obtain because we assume that the number and type of primitive driving patterns are unknown, thus we can not directly evaluate the learned model performance by comparing the learning results to the ground truth. Instead, we evaluate the utility of the developed approaches to segment different car-following driving data sequences into primitive driving patterns based on the ability of the learned models to predict the duration of each primitive driving pattern for the test data. Similar to \cite{taniguchi2015unsupervised}, the predictive log-likelihood is employed to evaluate the fidelity of the learned models. For each driver and corresponding test datasets, we apply the training data to learn parameters of all three models, and then the learned models are used to predict the probability distribution of durations for each primitive driving pattern at each frame of the test data. The probability distribution of durations of primitive driving patterns is used to compute the predictive log-likelihood by following \cite{taniguchi2015unsupervised}.  For each driver, the driving data are randomly and evenly grouped into ten folds. Nine folds are used for model training and the remaining one fold is used for model test. Thus we can finally obtain ten cross-validation results for each driver. During the training and testing procedures, the normalized car-following on-road data, $ \bar{\boldsymbol{x}} $, are used to learn the three model parameters and test these models.

\begin{table*}[t]
	\centering
	\caption{Statistical Results of Durations of Primitive Driving Patterns Using Three Methods for All Driving Data}
	\begin{tabular}{c|c|cccccccc}
		\hline
		\hline
		\multirow{2}{*}{Methods}& \multirow{2}{*}{Total Number} & \multicolumn{8}{c}{Duration of primitive driving pattern (s)}  \\ \cline{3-10}
	              &  &$ < $ 1.0 & [1.0, 5.0) & [5.0, 10.0) & [10.0, 15.0) & [15.0, 20.0) & [20.0, 25.0) & [25.0, 30.0) &$ \geqslant $ 30.0 \\
		\hline
		HDP-HMM & 18,174 & \cellcolor{lightgray!50} 0.0441 & 0.2819 & 0.4226 & 0.1534 & 0.0560 & 0.0228 & 0.0091 & 0.0102\\
		sticky HDP-HMM & \cellcolor{orange!90} 26,373 & \cellcolor{lightgray!40} 0.0216 & \textbf{0.5478} & \textbf{0.3276} & 0.0671 & 0.0213 & 0.0085 & 0.0032 & 0.0028 \\
		HDP-HSMM & \cellcolor{orange!90} 24,591 & \cellcolor{lightgray!30} 0.0065 & \textbf{0.4713} & \textbf{0.4211} & 0.0714 & 0.0193 & 0.0056 & 0.0018 & 0.0030 \\
		\hline
		\hline
	\end{tabular}
	\label{Table:statistical_segment}
\end{table*}

\subsubsection{Results}

Fig. \ref{figure:Method_Comparison} shows the log-likelihood of modeling the training data and the predictive log-likelihoods of the test data with the LOO-CV procedure,  given the training data for all three methods. Experiment results show that the HDP-HSMM obtains the largest log-likelihood of fitting training data (Fig. \ref{figure:Method_Comparison}, left) and the largest predictive log-likelihood of test data (Fig. \ref{figure:Method_Comparison}, right), indicating that the HDP-HSMM outperforms HDP-HMM for segmenting drivers' car-following data sequences. Note that the HDP-HSMM only obtain a slightly higher log-likelihood value and lower stand deviations, compared to the sticky HDP-HMM. 

\subsection{Segmentation Results and Comparisons}
Here, for clarity and conciseness, we only show the results from representative trails of driver \#0.  An example of segmentation results using the three approaches is shown in Fig. \ref{figure:seg_results1}. In order to demonstrate the advantages of HDP-HSMM, we make a further discussion and analysis as follows.

From the top plot in Fig. \ref{figure:seg_results1}, it can be noticed that the HDP-HMM could not segment the driving patterns as expected. For example, the driving data ranging from 0 s to 20 s obviously include the positive and negtive values of range rate and acceleration (i.e., the driver \#0, at least,  presents the closing in and falling behind patterns), but the HDP-HMM does treat the behaviors as one primitive pattern, instead of obtaining the expected segmentations to capture the underlying patterns.

Regarding sticky HDP-HMM (Fig. \ref{figure:seg_results1}, middle), it can segment driving data into different primitive patterns, but being sensitive to data fluctuation. For example, such patterns with very short time duration occur frequently (e.g., a pattern only keeps about 0.2 s starting at 8.4 s), which is not expected for drivers' driving states. The sticky HDP-HMM results in many such patterns whose duration is less than 1.0 s, for example, durations are 0.2 s, 0.8 s, 0.1 s, 0.7 s, 0.3 s, 0.6 s  at time 8.6 s, 32.2 s, 55.9 s, 58.8 s, 64.3 s and 64.7 s, respectively. In real driving cases, human drivers obviously do not adjust their driving modes or primitive driving patterns with a high frequency \cite{yang2010development} such as only keeping acceleration 0.2 s and then taking deceleration in normal driving.

\begin{figure}[t]
	\centering
	\includegraphics[width = 0.48\textwidth]{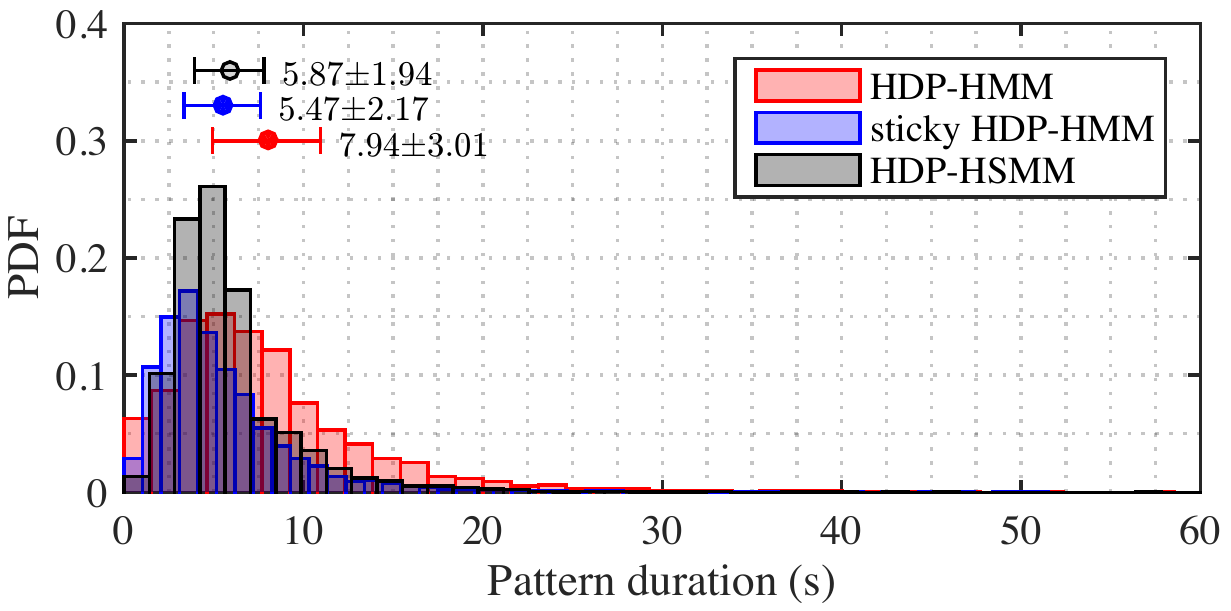}
	\caption{Statistical results of primitive driving pattern durations.}
	\label{fig:StatisticalDuration}
\end{figure}

\begin{figure}[ht!]
	\centering
	\subfloat[HDP-HMM]{\includegraphics[width = 0.48\textwidth]{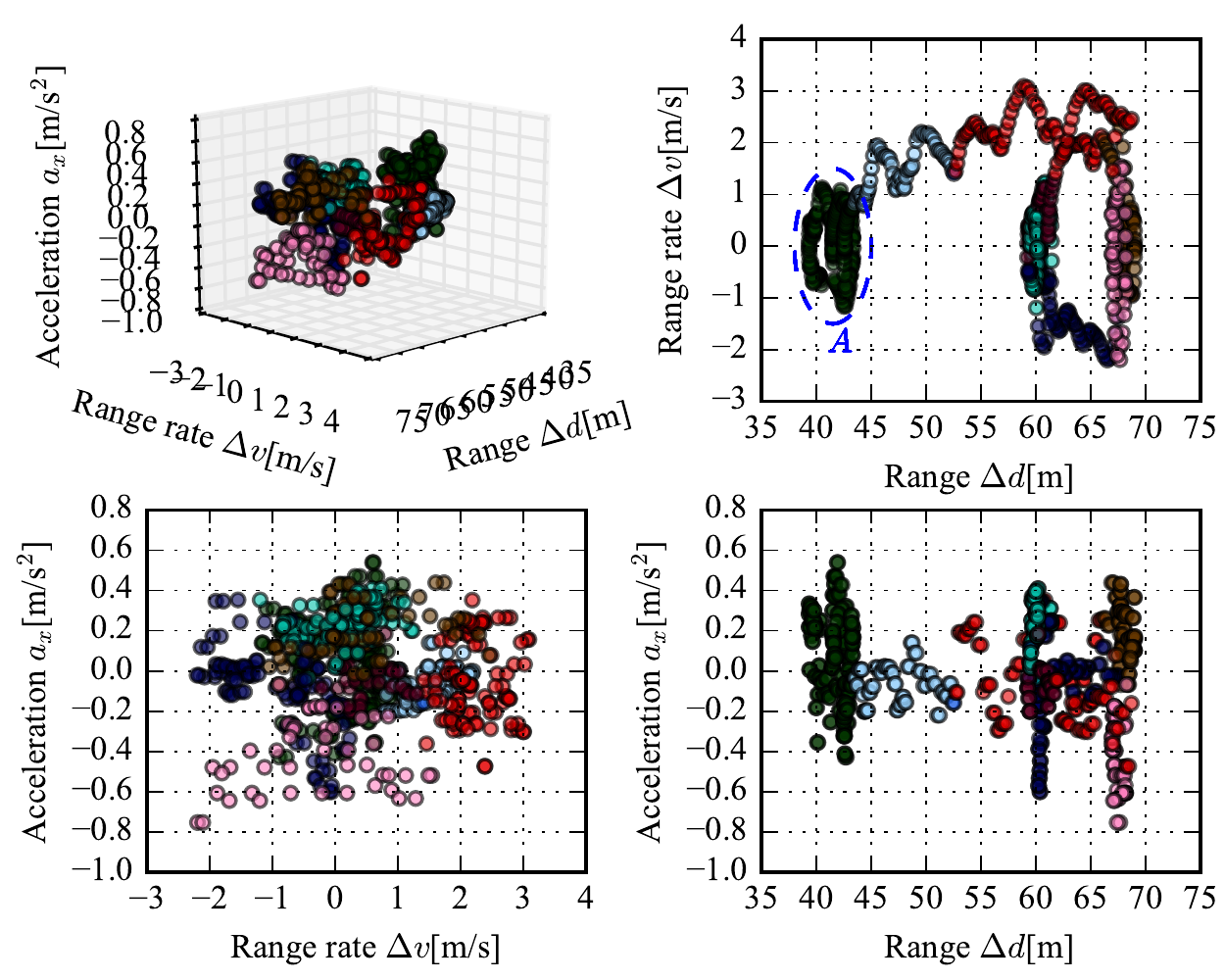}}\\
	\subfloat[sticky HDP-HMM]{\includegraphics[width = 0.48\textwidth]{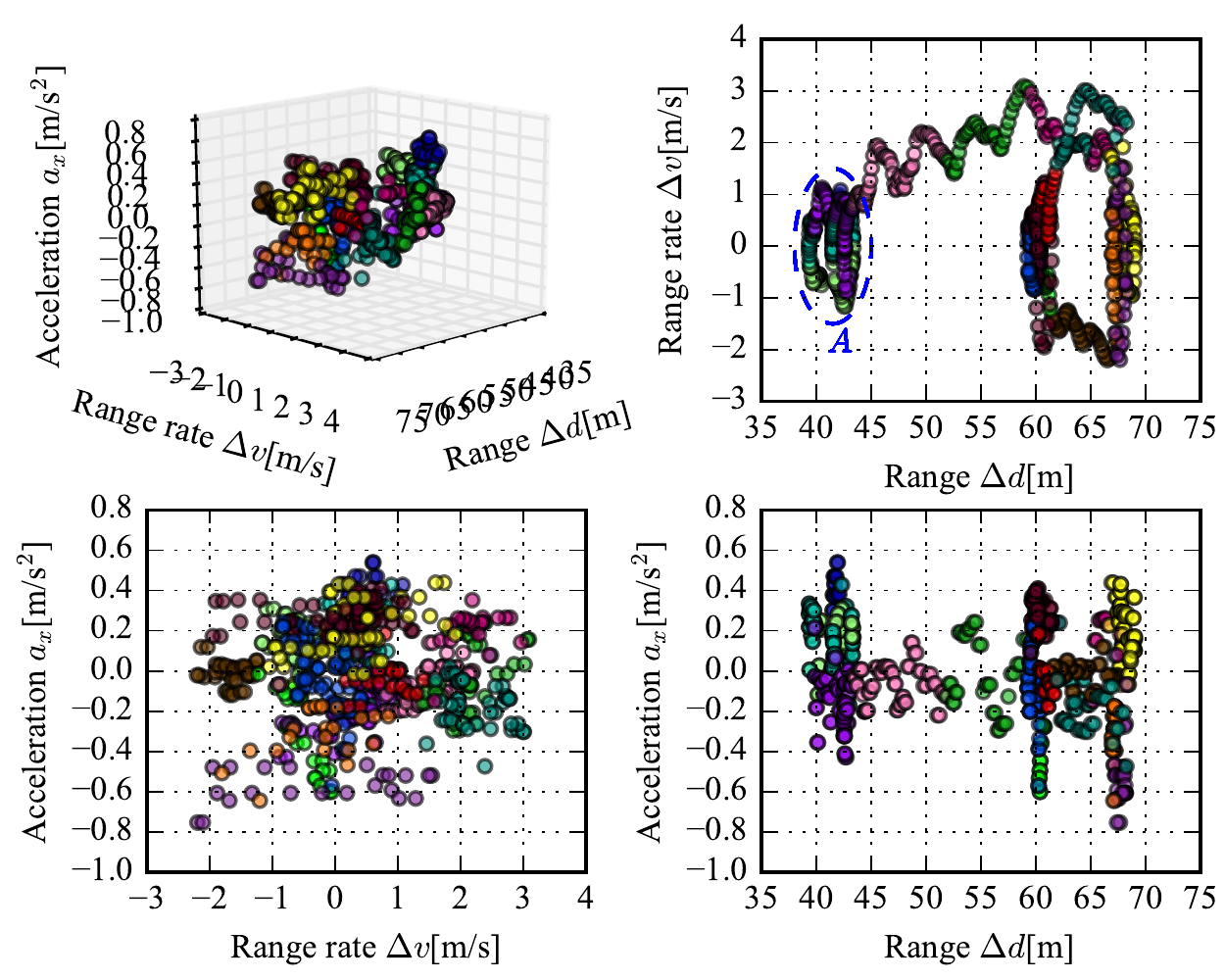}}\\
	\subfloat[HDP-HSMM]{\includegraphics[width = 0.48\textwidth]{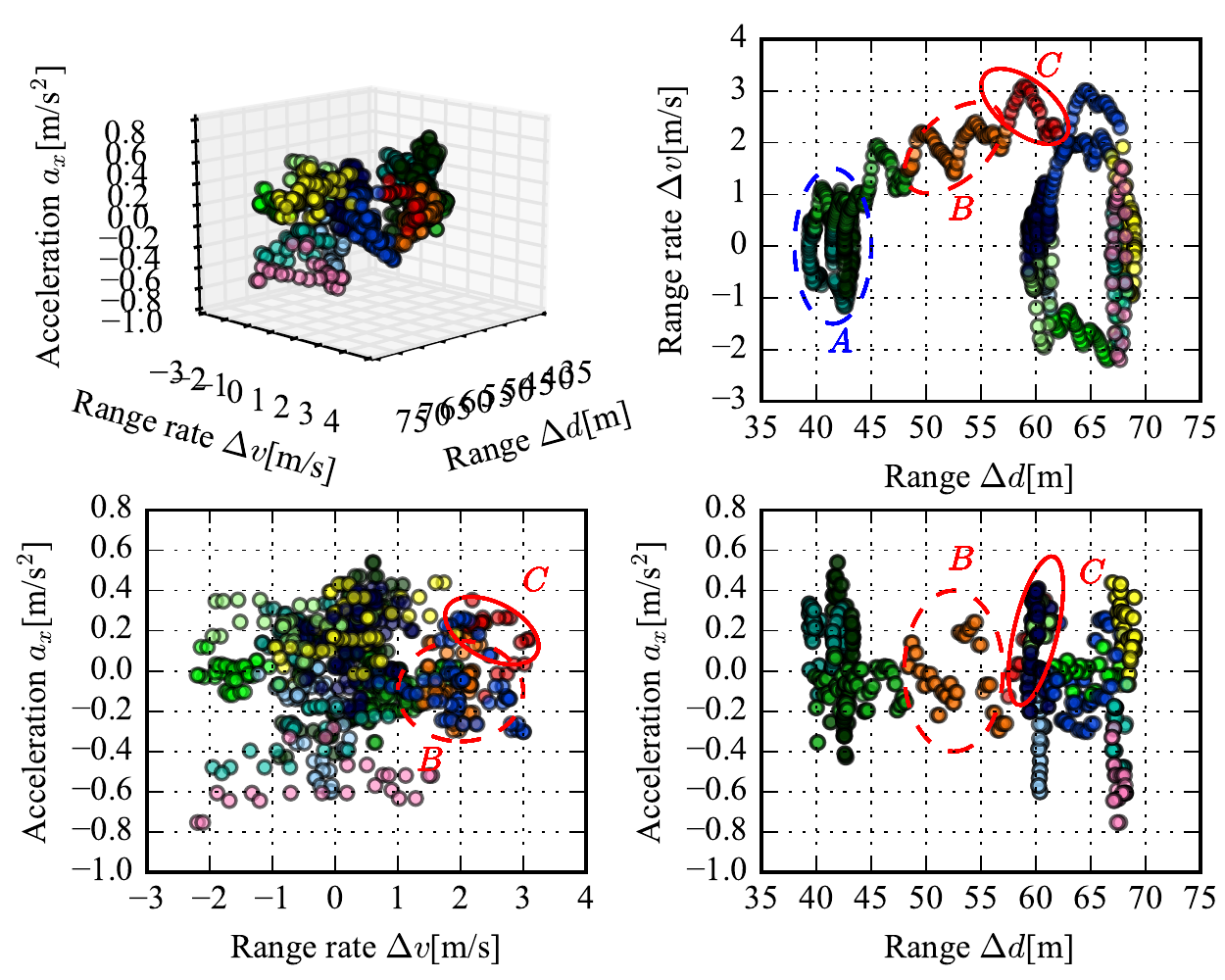}}
	\caption{Segment results with one event for driver \#0 using three methods.}
	\label{figure:seg_results2}
\end{figure}

\begin{figure}[t]
	\centering
	\includegraphics[width = 0.48\textwidth]{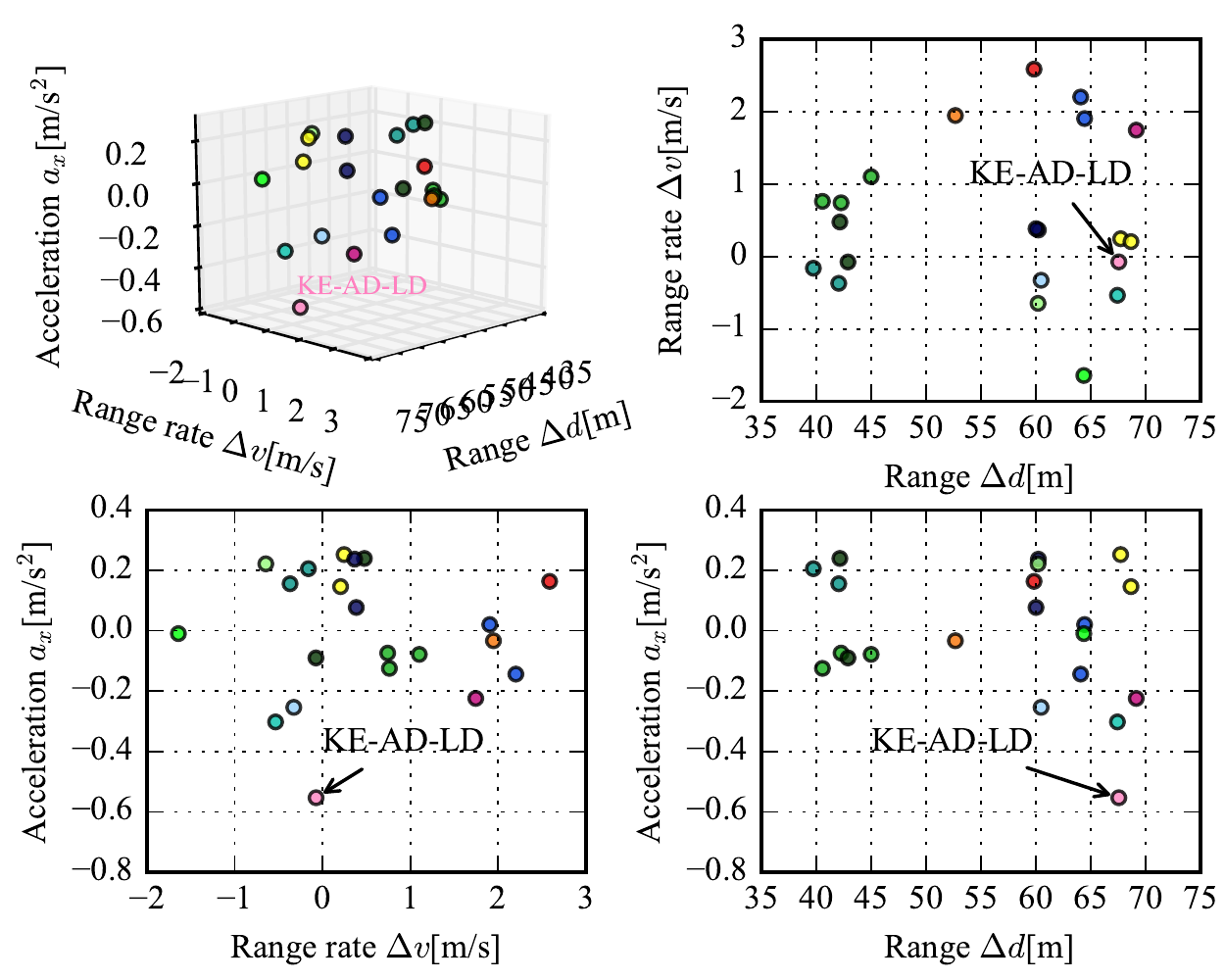}
	\caption{Example of clustering results for driver \#0 using K-means clustering method based on the HDP-HSMM with K = 1, where KE-AD-LD represents ``keeping (KE) distance by aggressive deceleration (AD) in a long distance (LD)''.}
	\label{figure:cluster_res}
\end{figure}

\begin{figure*}[t]
	\centering
	\subfloat[76 events for driver \#0]{\includegraphics[width = 0.24\textwidth]{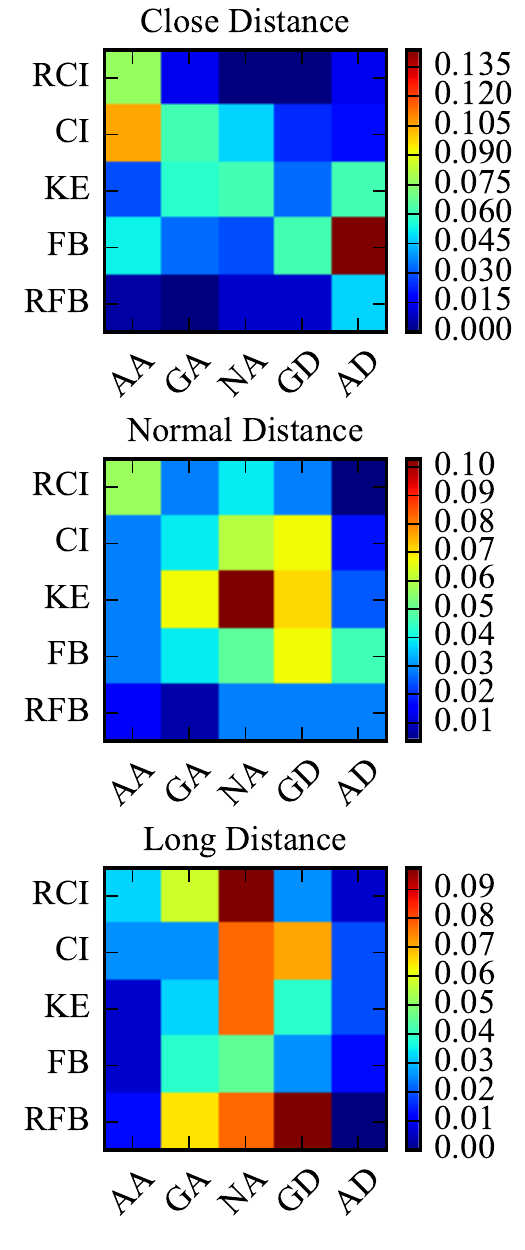}}
    \subfloat[92 events for driver \#1]{\includegraphics[width = 0.24\textwidth]{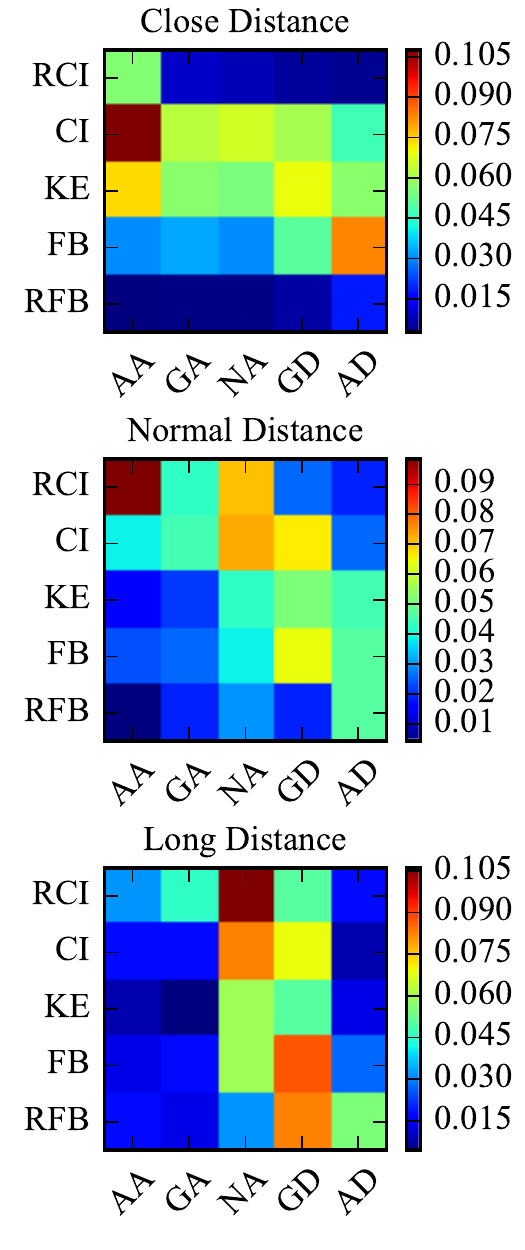}}
    \subfloat[96 events for driver \#2]{\includegraphics[width = 0.24\textwidth]{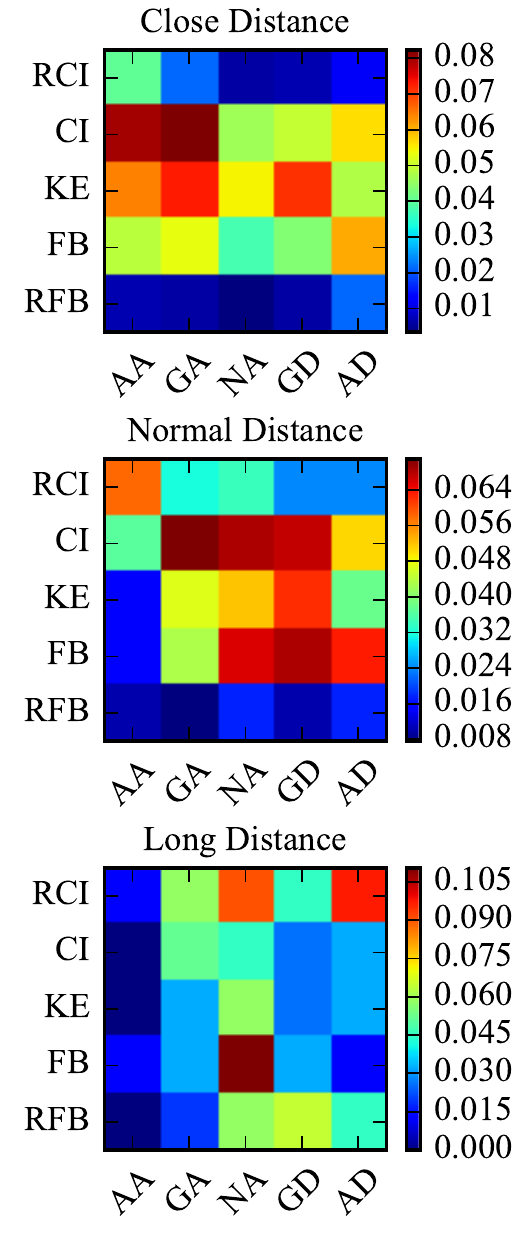}}
  \subfloat[111 events for driver \#3]{\includegraphics[width = 0.24\textwidth]{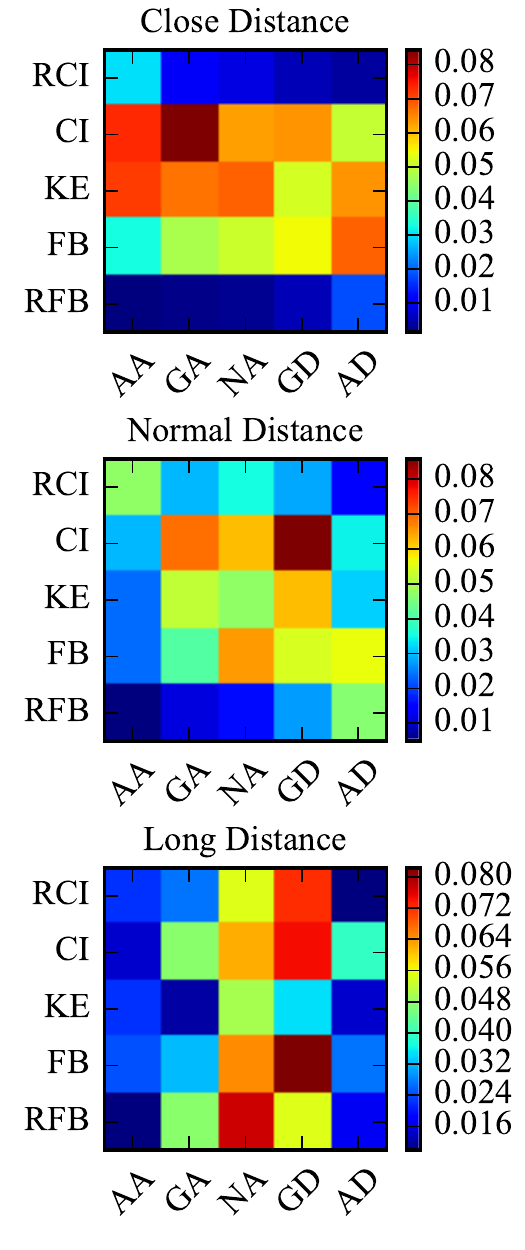}}
	\caption{Normalized frequency distributions of driving patterns for four drivers using the HDP-HSMM method. The dark red and dark blue indicate a higher and lower probability of driving in this pattern, respectively. RCI = rapidly closing in, CI = closing in, KE = keeping, FB = falling behind, RFB = rapidly falling behind, AA = aggressive acceleration, GA = gentle acceleration, NA = no acceleration, GD = gentle deceleration, AD = aggressive deceleration.}
	\label{figure:driving_style}
\end{figure*}

Table \ref{Table:statistical_segment} presents the statistical results of the durations of primitive driving patterns for all driving data using the three approaches. We found that the HDP-HSMM obtains a similar amount of the primitive driving patterns with the sticky HDP-HMM approach, as highlighted in orange. In addition, the HDP-HSMM approach obtains the lowest percentage (with 0.0065) of primitive driving pattern whose duration is less than 1.0 s, compared to HDP-HMM with percentage of 0.0216 and sticky HDP-HMM with percentage of 0.0441, as shaded with gray. It indicates that the HDP-HSMM method can segment time series driving data into several segments and hold the primitive driving patterns to be an expectable duration.  Therefore, in the following section, driving styles will be analyzed and evaluated based on results of HDP-HSMM. 

Fig. \ref{fig:StatisticalDuration} presents the statistical results including means and standard deviations of pattern durations using the three approaches. We can see that most driving primitive patterns are approximately falling in 3.93 s - 7.81 s with HDP-HSMM and in 3.30 s - 7.64 s with sticky HDP-HMM. However, the HDP-HMM obtains a higher primitive pattern duration with 7.94 s on average, compared to the other two approaches with 5.47 s and 5.87 s on average. According to the segmentation results using HDP-HSMM and sticky HDP-HMM, therefore, we can conclude that human drivers usually keep a specific following mode about 3.50 s - 10.0 s when following a lead vehicle. In addition, note that drivers sometimes can stay in specific but rare primitive driving patterns with durations of more than 20.0 s when following a car. 

Fig. \ref{figure:seg_results2} shows the scattering results of primitive segments using all three methods. We can know that the HDP-HSMM (Fig. \ref{figure:seg_results2}, bottom) can segment time series data into a set of reasonable patterns with lower frequency of switching between driving states, compared to other two approaches. For example, in the region \textit{A} of Fig. \ref{figure:seg_results2}, the HDP-HMM can not recognize the underlying patterns. In addition, the sticky HDP-HMM method is able to characterize the latent patterns, but it generates patterns with a high changing frequency between patterns and short time durations, which is not consistent with that in the real driving case\cite{yang2010development}.

\subsection{Labeling Behavioral Semantics}

Based on the results of the learned primitive driving patterns using HDP-HSMM, drivers' car-following behaviors can be described by behavioral semantics defined as in Table \ref{Table:Var_Sem}. Here, in order to capture the dynamic process of behavior, we describe each primitive driving pattern in car-following data sequence using the following interpretation:
\begin{equation*}
\mathrm{ The \ driver \ is \ S}_{\Delta v} \ \mathrm{the \ lead \ vehicle \ by \ } \mathrm{S}_{a_{x}} \mathrm{ \ in \ a \ S}_{\Delta d} 
\end{equation*}
where 
\begin{equation*}
\begin{split}
&\mathrm{S}_{\Delta v} \in \{\mathrm{RCI}, \mathrm{CI}, \mathrm{KE}, \mathrm{FB}, \mathrm{RFB} \}; \\
&\mathrm{S}_{a_{x}} \in \{ \mathrm{AA}, \mathrm{GA}, \mathrm{NA}, \mathrm{GD}, \mathrm{AD}\}; \\
&\mathrm{S}_{\Delta d} \in \{ \mathrm{LD}, \mathrm{ND}, \mathrm{CD} \}.
\end{split}
\end{equation*}

For example, regarding the region \textit{B} of bottom plot in Fig. \ref{figure:seg_results2}, the relative range rate $ \Delta v > $ 1.0 m/s (i.e., $ \{\mathrm{S}_{\Delta v} = \mathrm{RFB}\} $),  the range $ \Delta d \in $ [47, 57] m (i.e., $ \mathrm{S}_{\Delta d} = \mathrm{ND} $), and the absolute values of the acceleration mostly fall in [0.05, 0.24] (i.e., $ \mathrm{S}_{a_{x}} = \mathrm{GA} $ or $ \mathrm{S}_{a_{x}} = \mathrm{GD} $). Therefore, we can describe the driving behavior data in region \textit{B} as ``The driver is \textit{rapidly falling behind} the lead vehicle by \textit{gentile accelerating} or \textit{decelerating} in a \textit{normal distance}.'' Similarly, the driving behavior in region \textit{C} can be described as ``The driver is \textit{rapidly falling behind} the lead vehicle by \textit{gentile accelerating} in a \textit{long distance}.'' In the following section, to be simple, we use a primitive pattern sequence to represent drivers' behavioral semantics such as [$ \mathrm{S}_{\Delta v},\mathrm{S}_{a_x},\mathrm{S}_{\Delta d} $]. 

Based on the learned results, some primitive driving patterns are easy to be labeled with semantic elements, but some are not. In order to make patterns be easier labeled and find the commons in primitive patterns of car-following behaviors, we cluster the driving data of each segment into a point to represent this primitive driving pattern. The K-means clustering method is applied in this work because it has been widely used to label driving patterns \cite{higgs2015segmentation,wang2016rapid,wang2015online} and shown its advantages. The clustering results allow us to tractably translate each primitive driving pattern in semantics. For example, Fig. \ref{figure:cluster_res} presents a clustering result for driver \#0 based on the segmentation results of HDP-HSMM. According to Table \ref{Table:Var_Sem}, the pink point can be semantically interpreted as [$ \mathrm{KE},\mathrm{AD},\mathrm{LD} $], that is, ``The driver is \textit{keeping} distance with the lead vehicle by \textit{aggressive deceleration} in a \textit{long distance}.'', corresponding to the pink points in the bottom plot of Fig. \ref{figure:seg_results2}.

\section{Driving Style Analysis Based on Primitive Driving Patterns}
Understanding drivers' behavioral semantics is able to facilitate and help driving style analysis. According to the above discussion, we know that the HDP-HSMM provides a better semantical way for analyzing drivers' car-following behaviors. Therefore, in what follows, we will discuss driving styles using primitive driving patterns derived from HDP-HSMM.

\subsection{Normalized Frequency Distribution of Driving Patterns}
Instead of using the statistical metrics such as mean and standard deviation, we utilize the normalized frequency distribution of primitive driving patterns to characterize driving styles, which allows one to intuitively analyze driving habits. For each driver with $ L $ car-following events  $ X_{m} = \{\boldsymbol{x}_{m}^{(l)}\}_{l=1}^{L}$, the normalized probability of each pattern is computed by

\begin{equation}\label{eq:nfd}
f_{\star, \ast}^{(m)} = \frac{N^{(m)}_{\star, \ast}}{\sum_{\star} N^{(m)}_{\star, \ast}}, \ \ \ \star = [\mathrm{S}_{\Delta v},\mathrm{S}_{a_x}], \ \ \ \ast = \mathrm{S}_{\Delta d}
\end{equation}
where $ N^{(m)}_{\star, \ast} $ is the number of occurrences for driving pattern $ \star $  in $ X_{m} $ based on the distance pattern $ \ast $. Thus, we obtain the normalized frequency distribution for each driver with three distance patterns (i.e., close distance, normal distance, and long distance).  Each primitive driving pattern is clustered and labeled according to Table \ref{Table:Var_Sem}. 

\begin{figure}[t]
	\centering
	\subfloat[]{\begin{tikzpicture}[thick]
		% top right
		\fill[blue!90] (1.25,1.25) -- (2.5,1.25) -- (2.5,2.5) -- (1.25,2.5) -- cycle;
		\fill[fill = red, path fading=south, fading angle=-45]	(1.25,1.25) -- (2.5,1.25) -- (2.5,2.5) -- (1.25,2.5) -- cycle;
		
		% top left
		\fill[blue!90] (0,1.25) -- (1.25,1.25) -- (1.25,2.5) -- (0,2.5) -- cycle;
		\fill[fill = red, path fading=south, fading angle=45]	(0,1.25) -- (1.25,1.25) -- (1.25,2.5) -- (0,2.5) -- cycle;
		
		% bottom left
		\fill[blue!90] (0,0) -- (1.25,0) -- (1.25,1.25) -- (0,1.25) -- cycle;
		\fill[fill = red, path fading=south, fading angle=135]	(0,0) -- (1.25,0) -- (1.25,1.25) -- (0,1.25) -- cycle;
		
		% bottom right 
		\fill[blue!90] (1.25,0) -- (2.5,0) -- (2.5,1.25) -- (1.25,1.25) -- cycle;
		\fill[fill = red, path fading=south, fading angle=-135]	(1.25,0) -- (2.5,0) -- (2.5,1.25) -- (1.25,1.25) -- cycle;
		
		% draw grid 
		\draw[step=0.5] (0,0) grid (2.5,2.5);
		
		% add labels
		% x-axis
		\node at (0.15,-0.25) {$ -2 $};
		\node at (0.75,-0.25) {$ -1 $};
		\node at (1.3,-0.25) {$ 0 $};
		\node at (1.75,-0.25) {$ 1 $};
		\node at (2.25,-0.25) {$ 2 $};
		
		% y-axis
		\node at (-0.25,0.25) {$ 2 $};
		\node at (-0.25,0.75) {$ 1 $};
		\node at (-0.25,1.25) {$ 0 $};
		\node at (-0.4,1.75) {$ -1 $};
		\node at (-0.4,2.25) {$ -2 $};
		
		% note and arrow
		\node at (3, 0.25) {$ \mathrm{S}_{a_{x}} $};
		\node at (0.33, 3) {$ \mathrm{S}_{\Delta v} $};
		% arrow
		\draw[->] (0,0) -- (3.3,0);
		\draw[->] (0,0) -- (0,3.3);
		
		\end{tikzpicture}}
	\subfloat[]{\begin{tikzpicture}[thick]
		\draw[step=0.5] (0,0) grid (2.5,2.5);
		\draw[fill = green] (1.5,1.5) -- (2,1.5) -- (2,2) -- (1.5,2) -- cycle;
		\node at (1.75,1.75) {$ i,j $};
		\node at (1.75,-0.25) {$ i $};
		\node at (-0.25,1.75) {$ j $};
		
		% note and arrow
		\node at (3, 0.25) {$ \mathrm{S}_{a_{x}} $};
		\node at (0.33, 3) {$ \mathrm{S}_{\Delta v} $};
		% arrow
		\draw[->] (0,0) -- (3.3,0);
		\draw[->] (0,0) -- (0,3.3);
		\end{tikzpicture}}
	\caption{The schematic diagram of (a) driving style patterns and (b) pattern index ($ i,j $) for pattern [$  \mathrm{S}_{a_{x}},\mathrm{S}_{\Delta v} $].}
	\label{figure:driving_patterns}
\end{figure}
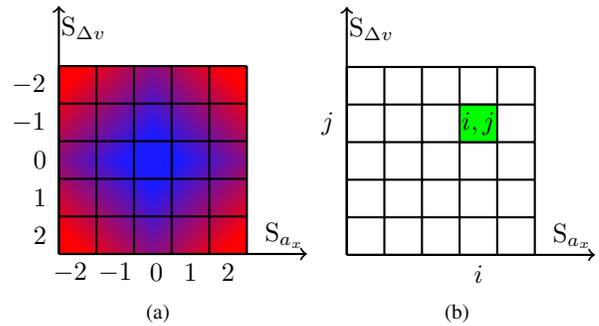

Fig. \ref{figure:driving_style} shows examples of the normalized frequency distribution of primitive driving patterns for four drivers. Dark red represents that the driver has a higher probability of acting in this pattern and dark blue represents that the driver has a lower probability (nearly equal to zero) of driving in this pattern. For instance, when following a lead vehicle in a long distance (Fig. \ref{figure:driving_style}, bottom), driver \#0 (Fig. \ref{figure:driving_style}(a)) and driver \#1 (Fig. \ref{figure:driving_style}(b)) prefer to rapidly close in the lead vehicle by no acceleration/deceleration, while driver \#2 (Fig. \ref{figure:driving_style}(c)) and driver \#3 (Fig. \ref{figure:driving_style}(d)) prefer falling behind by no acceleration/deceleration and gentle deceleration, respectively. When following the lead vehicle in a close or normal distance, our proposed approaches can also provide an intuitive explanation for researchers.

In order to show which primitive pattern drivers much prefer, we assign an unique index value to each primitive driving pattern in car-following, as shown in Fig. \ref{figure:driving_patterns}. Acceleration and range rate with five levels are both labeled using integers ranging from -2 to 2. A larger absolute value of the index indicates a more aggressive driving style, represented by dark red; conversely, a smaller absolute value of the index represents a more gentle driving style, represented by dark blue, as shown in Fig. \ref{figure:driving_patterns}(a). The pattern with the largest normalized frequency probability in each distance pattern for driver $ m $ is treated as the one this person prefers, represented using its index $ i $ and $ j $ (Fig. \ref{figure:driving_patterns}(b)). For example, when following lead vehicles with a close distance, the primitive driving pattern that driver \#0 strongly prefers is [$ \mathrm{FB} $, $ \mathrm{AD} $], denoted as the darkest red in the top plot of Fig. \ref{figure:driving_style}(a). Table \ref{Table:driver_prefer} lists the index value ($ i,j $) of the most preferring primitive driving patterns for all drivers, where $ i $ and $ j $ represent the levels of acceleration/deceleration and range rate, respectively. According to Table \ref{Table:driver_prefer}, we can conclude:
\begin{itemize}
	\item When following a lead vehicle in a close distance, most drivers prefer falling behind or closing in (i.e., $ j = -1 $ or $ 1 $) the lead vehicle by taking an aggressive acceleration or deceleration (i.e., $ i = -2$ or $ 2 $), except driver \#11 (blue shade) who prefers to keep a constant headway distance by no acceleration/deceleration. In addition, driver \#4, driver \#6, and driver \#17 (red shade) prefer to rapidly fall behind  (i.e., $ j = -2 $)  the lead vehicle by accelerating aggressively (i.e., $ i = -2 $).
	
	\item When following a lead vehicle with a normal distance, most drivers prefer falling behind or closing in (i.e., $ j = 1 $ or $ 1 $) the lead vehicle by taking a gentle deceleration (i.e., $ i = 1$), except driver \#1, driver \#4, and driver \#17 (orange shade) who prefer rapidly closing in the lead vehicle (i.e., $ j = -2 $) by taking an aggressive acceleration (i.e., $ i = -2 $).
	
	\item When following a lead vehicle with a long distance, most drivers prefer closing in (i.e., $ j =1 $) with no acceleration/deceleration (i.e., $ i = 0 $), except driver \#0, driver \#2, driver \#6, driver \#12, and driver \#15 noted with light gray shade. For example, driver \#12 prefers falling behind by taking an aggressive deceleration.
\end{itemize}

Therefore, our proposed framework for driving styles analysis based on primitive driving patterns can semantically analyze individual driving styles, instead of only using the statistical metrics such as means and standard deviations.

\begin{table}[t]
	\centering
	\caption{\textsc{Driver's Preference in Driving Primitive Pattern ($ i,j $) for All Participants.}}
	\begin{tabular}{c|c|c|c}
		\hline
		\hline
		Driver No. & CD ($ i,j $) & ND ($ i,j $) & LD ($ i,j $) \\
		\hline
		0 & (2,1)                                          & (0,0)                                  & \cellcolor{lightgray!40}(1,2) or (0,-2) \\
		1 & (-2,-1)                                      & \cellcolor{orange!90}(-2,-2) & \cellcolor{lightgray!40}(0,-2) \\
		2 & (-1,-1)                                      & (-1,-1)                               & (0,1) \\
		3 & (-1,-1)                                      & (1,-1)                                 & (1,1) \\
		4 & \cellcolor{orange!1!red}(-2,-2)    & \cellcolor{orange!90}(-2,-2) & (0,1) \\
		5 & (-2,-1)                                      & (0,-1)                                 & (0,-1) \\
		6 & \cellcolor{orange!1!red}(-2,-2)    & (2,-1)                                  & \cellcolor{lightgray!40}(1,-2) \\
		7 & (2,1)                                          & (0,1)                                   & (0,1) \\
		8 & (-2,0)                                        & (1,1)                                   & (0,1) \\
		9 & (2,1)                                          & (1,-1)                                  & (0,-1) \\
		10 & (-2,-1)                                     & (1,0)                                   & (0,-2) \\
		11 & \cellcolor{orange!1!blue!60}(0,0) & \cellcolor{orange!1!blue!60}(0,0) & \cellcolor{orange!1!blue!60}(0,1) \\
		12 & (2,1)                                        & (1,1)                                    & \cellcolor{lightgray!40}(2,1) \\
		13 & (-2,-1)                                    & (1,1)                                     & (0,1) \\
		14 & (-2,-1)                                    & (0,-1)                                   & (0,-1) \\
		15 & (2,1)                                        & (1,1)                                    & \cellcolor{lightgray!40}(1,2) \\
		16 & (1,1)                                        & (1,-1)                                   & (0,1) \\
		17 & \cellcolor{orange!1!red}(-2,-2)  & \cellcolor{orange!90}(-2,-2)    & (1,-1) \\
		\hline
		\hline
	\end{tabular}
	\label{Table:driver_prefer}
\end{table}

\subsection{Interdriver Differences in Driving Style}

\begin{figure}[t]
	\centering
	\includegraphics[width = 0.48 \textwidth]{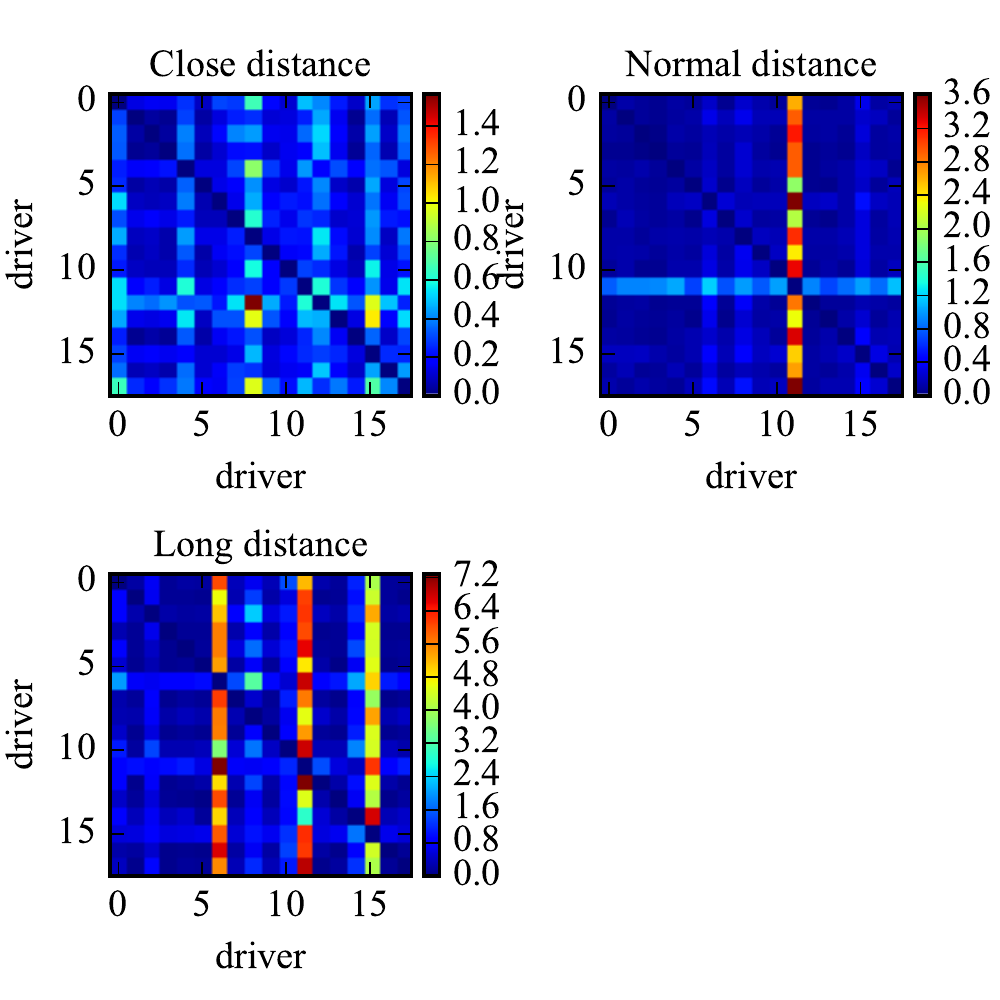}
	\caption{KL divergence between two normalized distributions of primitive driving patterns among all drivers.}
	\label{figure:KL_divergence}
\end{figure}

According to the above discussion, the normalized distribution of primitive driving pattern is able to describe and analyze driving styles for individuals. In this section, we will discuss the similarity or divergence in driving styles between drivers. Differing previous research using statistical metrics (e.g., mean and standard deviations) of driving data to capture driver's driver style, in this paper, we select the normalized distribution of primitive driving pattern as the indicator. When comparing the driving styles of two drivers, it is common to assume a normal distribution for the indicators and to compare their means and stand deviations\cite{lefevre2016learning}.  Here, we are not restricted by the limitations of this assumption and we compute the similarity of two drivers in driving styles using the Kullback-Leibler (KL) divergence\cite{mahboubi2016learning,wang2017much} between two corresponding distributions  to illustrate interdriver differences. The KL divergence of the normalized distribution of primitive driving patterns between drivers $ m_{i} $ and $ m_{j} $ is defined as

\begin{equation}
D_{KL}^{(\ast)}(F^{(m_i)}\parallel F^{(m_j)}) = -\sum_{\star}f^{(m_i)}_{\star,\ast} \log\left(  \frac{f^{(m_i)}_{\star,\ast}}{f^{(m_j)}_{\star,\ast}}\right) 
\end{equation}
where $ \star $ and $ \ast $ are defined same as in (\ref{eq:nfd}). A larger value of KL divergence indicates a big difference between two drivers. For example, when $ m_{i} = m_{j} $, meaning that the driver is compared to him/herself, thus we have $ D_{KL}^{(\ast)}(F^{(m_i)}\parallel F^{(m_j)}) = 0 $.

Fig. \ref{figure:KL_divergence} presents the KL divergence of each pairs of drivers. Dark red represents a great difference between two drivers and dark blue indicates that drivers are strongly similar to each other. From Fig. \ref{figure:KL_divergence}(a), we know that when driver follows a lead vehicle at a close distance, driver \#8 is different from others, especially different from driver \#12, i.e., the darkest red square. According to Fig. \ref{figure:KL_divergence}(b), when following the lead vehicle at a normal distance, driver \#11 is strongly different from others because there is a red vertical bar between driver \#11 and all other drivers. When following a long distance to the lead vehicle as shown in Fig. \ref{figure:KL_divergence}(c), driver \#6, driver \#11, and driver \#15 are strongly different from others. Therefore, the KL divergence index can provide a flexible way to illustrate the similarity or divergence in driving styles between drivers from a semantical perspective.

\section{Conclusion}
This paper presents a new framework for driving style analysis using primitive driving patterns with Bayesian nonparametric methods. First, a set of primitive driving patterns are learned from normalized naturalistic driving data by assuming the number of primitive driving patterns is unknown. To achieve this,  a hierarchical structure (i.e., HDP-HSMM) is developed by combining hierarchical Dirichlet process (HDP) and hidden semi-Markov model (HSMM). We also compare the HDP-HSMM method to other Bayesian nonparametric methods such as HDP-HMM and sticky HDP-HMM, and found that the proposed HDP-HSMM can learn a set of expected primitive driving patterns in car-following behaviors. To describe primitive driving pattern in semantics, this paper divides each feature variable into different levels according to drivers' physical and psychological perception thresholds. Instead of using a series of statistical results such mean and standard deviation, we use the normalized frequency distribution of primitive driving patterns to analyze individual driving style and also use the KL-divergence between distributions to illustrate differences between drivers.  Experimental results demonstrate that the primitive driving pattern-based analysis framework proposed in this paper provide an opportunity to model and analyze driving styles semantically.

% or


%\appendices
%% you can choose not to have a title for an appendix
%% if you want by leaving the argument blank
%\section{}

%\section*{Acknowledgment}
%
%
%The authors would like to thank..

\ifCLASSOPTIONcaptionsoff
  \newpage
\fi



\bibliographystyle{IEEEtran}
\bibliography{Lit_Ref}

\begin{thebibliography}{10}
\providecommand{\url}[1]{#1}
\csname url@samestyle\endcsname
\providecommand{\newblock}{\relax}
\providecommand{\bibinfo}[2]{#2}
\providecommand{\BIBentrySTDinterwordspacing}{\spaceskip=0pt\relax}
\providecommand{\BIBentryALTinterwordstretchfactor}{4}
\providecommand{\BIBentryALTinterwordspacing}{\spaceskip=\fontdimen2\font plus
\BIBentryALTinterwordstretchfactor\fontdimen3\font minus
  \fontdimen4\font\relax}
\providecommand{\BIBforeignlanguage}[2]{{%
\expandafter\ifx\csname l@#1\endcsname\relax
\typeout{** WARNING: IEEEtran.bst: No hyphenation pattern has been}%
\typeout{** loaded for the language `#1'. Using the pattern for}%
\typeout{** the default language instead.}%
\else
\language=\csname l@#1\endcsname
\fi
#2}}
\providecommand{\BIBdecl}{\relax}
\BIBdecl

\bibitem{di2014stochastic}
S.~Di~Cairano, D.~Bernardini, A.~Bemporad, and I.~V. Kolmanovsky, ``Stochastic
  {MPC} with learning for driver-predictive vehicle control and its application
  to {HEV} energy management,'' \emph{IEEE Transactions on Control Systems
  Technology}, vol.~22, no.~3, pp. 1018--1031, 2014.

\bibitem{sagberg2015review}
F.~Sagberg, Selpi, G.~F. Bianchi~Piccinini, and J.~Engstr{\"o}m, ``A review of
  research on driving styles and road safety,'' \emph{Human factors}, vol.~57,
  no.~7, pp. 1248--1275, 2015.

\bibitem{martinez17driving}
C.~M. Martinez, M.~Heucke, F.-Y. Wang, B.~Gao, and D.~Cao, ``Driving style
  recognition for intelligent vehicle control and advanced driver assistance: A
  survey,'' \emph{IEEE Transactions on Intelligent Transportation Systems},
  DOI:10.1109/TITS.2017.2706978, 2017.

\bibitem{rafael2006impact}
M.~Rafael, M.~Sanchez, V.~Mucino, J.~Cervantes, and A.~Lozano, ``Impact of
  driving styles on exhaust emissions and fuel economy from a heavy-duty truck:
  Laboratory tests,'' \emph{International Journal of Heavy Vehicle Systems},
  vol.~13, no. 1-2, pp. 56--73, 2006.

\bibitem{murphey2009driver}
Y.~L. Murphey, R.~Milton, and L.~Kiliaris, ``Driver's style classification
  using jerk analysis,'' in \emph{Computational Intelligence in Vehicles and
  Vehicular Systems, 2009. CIVVS'09. IEEE Workshop on}.\hskip 1em plus 0.5em
  minus 0.4em\relax IEEE, 2009, pp. 23--28.

\bibitem{xu2015establishing}
L.~Xu, J.~Hu, H.~Jiang, and W.~Meng, ``Establishing style-oriented driver
  models by imitating human driving behaviors,'' \emph{IEEE Transactions on
  Intelligent Transportation Systems}, vol.~16, no.~5, pp. 2522--2530, 2015.

\bibitem{shi2015evaluating}
B.~Shi, L.~Xu, J.~Hu, Y.~Tang, H.~Jiang, W.~Meng, and H.~Liu, ``Evaluating
  driving styles by normalizing driving behavior based on personalized driver
  modeling,'' \emph{IEEE Transactions on Systems, Man, and Cybernetics:
  Systems}, vol.~45, no.~12, pp. 1502--1508, 2015.

\bibitem{wang2017driving}
W.~Wang, J.~Xi, A.~Chong, and L.~Li, ``Driving style classification using a
  semi-supervised support vector machine,'' \emph{IEEE Transactions on
  Human-Machine Systems}, DOI:10.1109/THMS.2017.2736948, 2017.

\bibitem{vaitkus2014driving}
V.~Vaitkus, P.~Lengvenis, and G.~{\v{Z}}ylius, ``Driving style classification
  using long-term accelerometer information,'' in \emph{Methods and Models in
  Automation and Robotics (MMAR), 2014 19th International Conference On}.\hskip
  1em plus 0.5em minus 0.4em\relax IEEE, 2014, pp. 641--644.

\bibitem{higgs2015segmentation}
B.~Higgs and M.~Abbas, ``Segmentation and clustering of car-following behavior:
  Recognition of driving patterns,'' \emph{IEEE Transactions on Intelligent
  Transportation Systems}, vol.~16, no.~1, pp. 81--90, 2015.

\bibitem{okuda2013modeling}
H.~Okuda, N.~Ikami, T.~Suzuki, Y.~Tazaki, and K.~Takeda, ``Modeling and
  analysis of driving behavior based on a probability-weighted {ARX} model,''
  \emph{IEEE Transactions on Intelligent Transportation Systems}, vol.~14,
  no.~1, pp. 98--112, 2013.

\bibitem{sekizawa2007modeling}
S.~Sekizawa, S.~Inagaki, T.~Suzuki, S.~Hayakawa, N.~Tsuchida, T.~Tsuda, and
  H.~Fujinami, ``Modeling and recognition of driving behavior based on
  stochastic switched {ARX} model,'' \emph{IEEE Transactions on Intelligent
  Transportation Systems}, vol.~8, no.~4, pp. 593--606, 2007.

\bibitem{johnson2011driving}
D.~A. Johnson and M.~M. Trivedi, ``Driving style recognition using a smartphone
  as a sensor platform,'' in \emph{Intelligent Transportation Systems (ITSC),
  2011 14th International IEEE Conference on}.\hskip 1em plus 0.5em minus
  0.4em\relax IEEE, 2011, pp. 1609--1615.

\bibitem{macadam1998using}
C.~MacAdam, Z.~Bareket, P.~Fancher, and R.~Ervin, ``Using neural networks to
  identify driving style and headway control behavior of drivers,''
  \emph{Vehicle System Dynamics}, vol.~29, no.~S1, pp. 143--160, 1998.

\bibitem{do2017human}
Q.~H. Do, H.~Tehrani, S.~Mita, M.~Egawa, K.~Muto, and K.~Yoneda, ``Human
  drivers based active-passive model for automated lane change,'' \emph{IEEE
  Intelligent Transportation Systems Magazine}, vol.~9, no.~1, pp. 42--56,
  2017.

\bibitem{nilsson2016if}
J.~Nilsson, J.~Silvlin, M.~Brannstrom, E.~Coelingh, and J.~Fredriksson, ``If,
  when, and how to perform lane change maneuvers on highways,'' \emph{IEEE
  Intelligent Transportation Systems Magazine}, vol.~8, no.~4, pp. 68--78,
  2016.

\bibitem{nilsson2016lane}
J.~Nilsson, M.~Br{\"a}nnstr{\"o}m, E.~Coelingh, and J.~Fredriksson, ``Lane
  change maneuvers for automated vehicles,'' \emph{IEEE Transactions on
  Intelligent Transportation Systems}, 2016.

\bibitem{taylor2015method}
J.~Taylor, X.~Zhou, N.~M. Rouphail, and R.~J. Porter, ``Method for
  investigating intradriver heterogeneity using vehicle trajectory data: a
  dynamic time warping approach,'' \emph{Transportation Research Part B:
  Methodological}, vol.~73, pp. 59--80, 2015.

\bibitem{ma2007behavior}
X.~Ma and I.~Andreasson, ``Behavior measurement, analysis, and regime
  classification in car following,'' \emph{IEEE Transactions on Intelligent
  Transportation Systems}, vol.~8, no.~1, pp. 144--156, 2007.

\bibitem{hamada2016modeling}
R.~Hamada, T.~Kubo, K.~Ikeda, Z.~Zhang, T.~Shibata, T.~Bando, K.~Hitomi, and
  M.~Egawa, ``Modeling and prediction of driving behaviors using a
  nonparametric {B}ayesian method with {AR} models,'' \emph{IEEE Transactions
  on Intelligent Vehicles}, vol.~1, no.~2, pp. 131--138, 2016.

\bibitem{taniguchi2016sequence}
T.~Taniguchi, S.~Nagasaka, K.~Hitomi, N.~P. Chandrasiri, T.~Bando, and
  K.~Takenaka, ``Sequence prediction of driving behavior using double
  articulation analyzer,'' \emph{IEEE Transactions on Systems, Man, and
  Cybernetics: Systems}, vol.~46, no.~9, pp. 1300--1313, 2016.

\bibitem{taniguchi2015unsupervised}
T.~Taniguchi, S.~Nagasaka, K.~Hitomi, K.~Takenaka, and T.~Bando, ``Unsupervised
  hierarchical modeling of driving behavior and prediction of contextual
  changing points,'' \emph{IEEE Transactions on Intelligent Transportation
  Systems}, vol.~16, no.~4, pp. 1746--1760, 2015.

\bibitem{yu2010hidden}
S.-Z. Yu, ``Hidden semi-{M}arkov models,'' \emph{Artificial intelligence}, vol.
  174, no.~2, pp. 215--243, 2010.

\bibitem{rabiner1989tutorial}
L.~R. Rabiner, ``A tutorial on hidden {M}arkov models and selected applications
  in speech recognition,'' \emph{Proceedings of the IEEE}, vol.~77, no.~2, pp.
  257--286, 1989.

\bibitem{johnson2013bayesian}
M.~J. Johnson and A.~S. Willsky, ``Bayesian nonparametric hidden semi-{M}arkov
  models,'' \emph{Journal of Machine Learning Research}, vol.~14, no. Feb, pp.
  673--701, 2013.

\bibitem{teh2006hierarchical}
Y.~W. Teh, M.~I.~Jordan, M.~J.~Beal, and D.~M.~Blei, ``Hierarchical {D}irichlet
  processes,'' \emph{Journal of the American Statistical Association}, vol.
  101, no. 476, pp. 1566--1581, 2006.

\bibitem{fox2011bayesian}
E.~Fox, E.~B. Sudderth, M.~I. Jordan, and A.~S. Willsky, ``Bayesian
  nonparametric inference of switching dynamic linear models,'' \emph{IEEE
  Transactions on Signal Processing}, vol.~59, no.~4, pp. 1569--1585, 2011.

\bibitem{mahboubi2016learning}
Z.~Mahboubi and M.~J. Kochenderfer, ``Learning traffic patterns at small
  airports from flight tracks,'' \emph{IEEE Transactions on Intelligent
  Transportation Systems}, 2016.

\bibitem{ryden2008versus}
T.~Ryd{\'e}n \emph{et~al.}, ``{EM }versus {M}arkov chain {M}onte {C}arlo for
  estimation of hidden {M}arkov models: A computational perspective,''
  \emph{Bayesian Analysis}, vol.~3, no.~4, pp. 659--688, 2008.

\bibitem{wulsin2014modeling}
D.~F. Wulsin, E.~B. Fox, and B.~Litt, ``Modeling the complex dynamics and
  changing correlations of epileptic events,'' \emph{Artificial intelligence},
  vol. 216, pp. 55--75, 2014.

\bibitem{wang2017learning}
W.~Wang, D.~Zhao, J.~Xi, and W.~Han, ``A learning-based approach for lane
  departure warning systems with a personalized driver model,'' \emph{arXiv
  preprint arXiv:1702.01228}, 2017.

\bibitem{wang2017development}
W.~Wang, D.~Zhao, J.~Xi, D.~J. LeBlanc, and J.~K. Hedrick, ``Development and
  evaluation of two learning-based personalized driver models for car-following
  behaviors,'' in \emph{American Control Conference}, Seattle, WA, USA, May
  2017, pp. 1133--1138.

\bibitem{wang2017much}
W.~Wang, C.~Liu, and D.~Zhao, ``How much data is enough? a statistical approach
  with case study on longitudinal driving behavior,'' \emph{IEEE Transactions
  on Intelligent Vehicles}, DOI:10.1109/TIV.2017.2720459, 2017.

\bibitem{bishop2007pattern}
C.~M. Bishop, ``Pattern recognition and machine learning,'' \emph{Springer, New
  York}, 2007.

\bibitem{ahmed1999modeling}
K.~I. Ahmed, ``Modeling drivers' acceleration and lane changing behavior,''
  Ph.D. dissertation, Massachusetts Institute of Technology, 1999.

\bibitem{yang2010development}
H.-H. Yang and H.~Peng, ``Development of an errorable car-following driver
  model,'' \emph{Vehicle System Dynamics}, vol.~48, no.~6, pp. 751--773, 2010.

\bibitem{fischer2012evaluation}
M.~Fischer, L.~Eriksson, and K.~Oeltze, ``Evaluation of methods for measuring
  speed perception in a driving simulator,'' in \emph{Proceedings of Driving
  Simulation Conference Europe, edited by S. Espi{\'e}, A. Kemeny, and F.
  M{\'e}rienne}, 2012, pp. 214--229.

\bibitem{macadam2003understanding}
C.~C. Macadam, ``Understanding and modeling the human driver,'' \emph{Vehicle
  System Dynamics}, vol.~40, no. 1-3, pp. 101--134, 2003.

\bibitem{ioannou1994precursor}
P.~Ioannou \emph{et~al.}, ``Precursor systems analysis for automated highway
  systems: Volume {II}-lateral and longitudinal control,'' \emph{Center for
  Advanced Transportation Technologies, University of Southern California.
  Technical Report submitted to Federal Highway Administration}, 1994.

\bibitem{wang2016rapid}
W.~Wang and J.~Xi, ``A rapid pattern-recognition method for driving styles
  using clustering-based support vector machines,'' in \emph{American Control
  Conference (ACC), 2016}.\hskip 1em plus 0.5em minus 0.4em\relax IEEE, 2016,
  pp. 5270--5275.

\bibitem{wang2015online}
S.~Wang, Y.~Zhang, C.~Wu, F.~Darvas, and W.~A. Chaovalitwongse, ``Online
  prediction of driver distraction based on brain activity patterns,''
  \emph{IEEE Transactions on Intelligent Transportation Systems}, vol.~16,
  no.~1, pp. 136--150, 2015.

\bibitem{lefevre2016learning}
S.~Lef{\`e}vre, A.~Carvalho, and F.~Borrelli, ``A learning-based framework for
  velocity control in autonomous driving,'' \emph{IEEE Transactions on
  Automation Science and Engineering}, vol.~13, no.~1, pp. 32--42, 2016.

\end{thebibliography}


% \end{thebibliography}



%\begin{IEEEbiography}{Wenshuo Wang}
\begin{IEEEbiography}[{\includegraphics[width=1in,height=1.25in,clip,keepaspectratio]{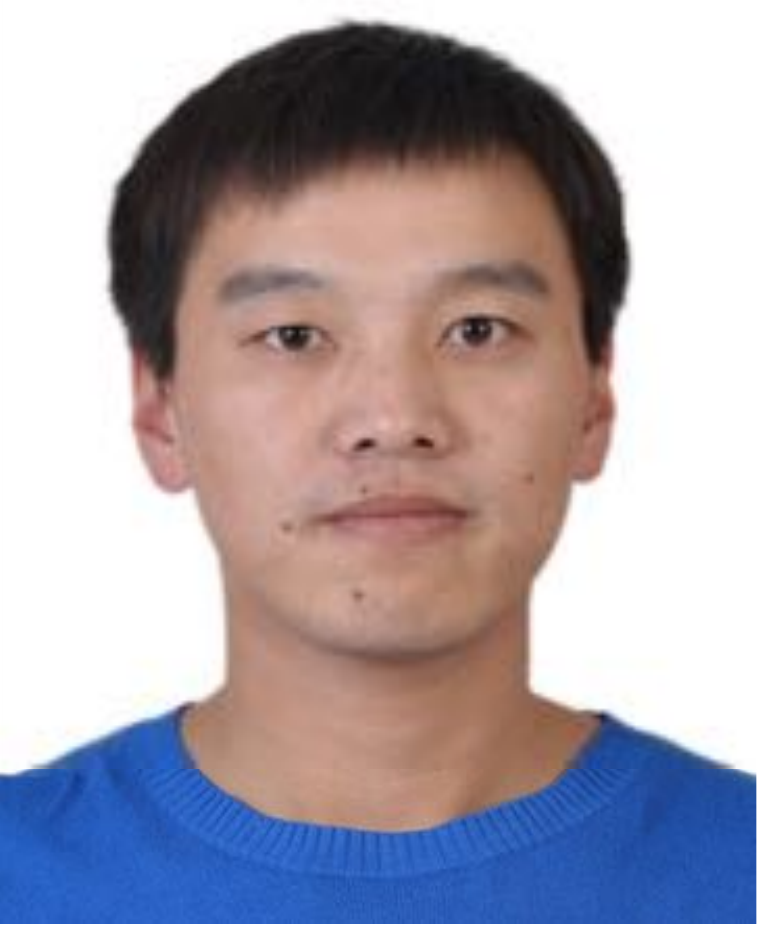}}]{Wenshuo Wang} 
	(S'15) received his B.S. in  Transportation Engineering from ShanDong University of Technology, Shandong, China, in 2012. He is a Ph.D. candidate for Mechanical Engineering, Beijing Institute of Technology (BIT). Now he is studying in the Vehicle Dynamics \& Control Lab, Department of Mechanical Engineering, University of California at Berkeley (UCB). His work focuses on modeling and recognizing drivers behavior, making intelligent control systems between human driver and vehicle.
\end{IEEEbiography}

\begin{IEEEbiography}[{\includegraphics[width=1in,height=1.25in,clip,keepaspectratio]{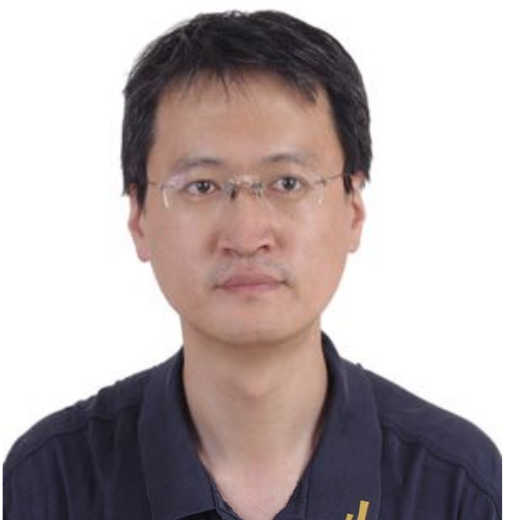}}]{Junqiang Xi}
	received the B.S. in Automotive Engineering from Harbin Institute of Technology, Harbin, China, in 1995 and the PhD in Vehicle Engineering from Beijing Institute of Technology (BIT), Beijing, China, in 2001. In 2001, he joined the State Key Laboratory of Vehicle Transmission, BIT. During 2012-2013, he made research as an advanced research scholar in Vehicle Dynamic and Control Laboratory, Ohio State University(OSU), USA. He is Professor and Director of Automotive Research Center in BIT currently. His research interests include vehicle dynamic and control, power-train control, mechanics, intelligent transportation system and intelligent vehicles.
\end{IEEEbiography}

\begin{IEEEbiography}[{\includegraphics[width=1in,height=1.25in,clip,keepaspectratio]{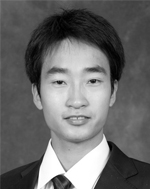}}]{Ding Zhao}
	received his Ph.D. degree in 2016 from the University of Michigan, Ann Arbor. He is currently an Assistant Research Scientist at Mechanical Engineering of the University of Michigan. His research interests include the autonomous vehicles, intelligent transportation, connected vehicles, dynamics and control, human-machine interaction, machine learning, and big data analysis.
\end{IEEEbiography}

%\begin{IEEEbiographynophoto}{Ding Zhao}

%\newpage

%\begin{IEEEbiographynophoto}{Junqiang Xi}


%\vfill


\end{document}